\documentclass[table, xcdraw]{sig-alternate-br} 
\usepackage[utf8]{inputenc}
\usepackage{fourier} 
\usepackage{array}
\usepackage{makecell}
\usepackage{pgfgantt}
\usepackage{xcolor}
\usepackage{balance} 
\usepackage{subfiles}
\usepackage{import}
\usepackage{appendix}
\usepackage{adjustbox}
\usepackage{caption}
\usepackage{subcaption}
\usepackage[a-1b]{pdfx}
\usepackage[utf8]{inputenc}
\usepackage{amsmath}
\usepackage{algorithm}
\usepackage{algpseudocode}
\usepackage{upgreek}
\usepackage{mathtools}
\usepackage{pgfplots}
\usepgfplotslibrary{external}
\usepackage[labelfont=bf, textfont=it]{caption}
\usepackage{booktabs,chemformula}
\usepackage{multirow}
\usepackage{graphicx} [table]
\usetikzlibrary{spy}

\begin{document}
\emergencystretch 3em

%
\conferenceinfo{33$^{rd}$ Twente Student Conference on IT}{July. 3$^{rd}$, 2020, Enschede, The Netherlands.}
\CopyrightYear{2020} 

\title{Activation function impact on Sparse Neural Networks}

\numberofauthors{1} 
\author{
\alignauthor
Adam Dubowski\\
       \affaddr{University of Twente}\\
       \affaddr{P.O. Box 217, 7500AE Enschede}\\
       \affaddr{The Netherlands}\\
       \email{a.dubowski@student.utwente.nl}
}
\date{3 May 2020}

\maketitle

\begin{abstract}
While the concept of a Sparse Neural Network has been researched for some time, researchers have only recently made notable progress in the matter. Techniques like Sparse Evolutionary Training allow for significantly lower computational complexity when compared to fully connected models by reducing redundant connections. That typically takes place in an iterative process of weight creation and removal during network training. Although there have been numerous approaches to optimize the redistribution of the removed weights, there seems to be little or no study on the effect of activation functions on the performance of the Sparse Networks. This research provides insights into the relationship between the activation function used and the network performance at various sparsity levels.

\end{abstract}


\section{Introduction}
Artificial Neural Network (ANN) is a powerful Deep Learning method inspired by the human brain architecture \cite{ohn2019smooth} which has been adapted for speech recognition, computer vision, natural language processing, and many others \cite{najafabadi2015deep}.\\
Although constantly researched and improved, ANNs still have many flaws. While most researchers strive to improve algorithms' accuracy, it is the efficiency of deep learning solutions that is still one of its major limitations. The most common implementation of ANNs is the Multilayer Perceptron (MLP) which consists of fully connected neuron layers. That causes high redundancy because most of the connections have little or no impact on the accuracy of the model and could be removed \cite{denil2013predicting}. One idea to improve the efficiency of ANNs is to limit the number of connections between layers, thus introducing sparsity in the model. Mocanu et al. suggested a training method called Sparse Evolutionary Training (SET) \cite{mocanu2018scalable} supposed to deliver significantly greater performance by cutting up to 99.9\% of connections \cite{liu2019sparse}.\\
Many other ideas on how to develop Sparse Neural Networks (SNNs) without training a fully connected model have been proposed, but most focus on the distribution of connections, while there has been little or no research on the impact of the choice of activation function used on the sparsity effect in the SNNs.
This research investigates this relation to understand whether the activation functions currently used for densely connected networks \cite{ohn2019smooth} still behave reliably in the sparse context and which functions shall be recommended for SNN implementations. The findings will hopefully help researchers and developers make better decisions while choosing activation functions and sparsity levels for their models, thus improving the performance of their models. The research questions answered  in the paper are:\\
\textbf{RQ1} What is the impact of activation functions on the sparsity sweep and accuracy of SNNs?\\
\textbf{RQ2} Is there a single activation function to be preferred in SNN implementations?\\
\textbf{RQ3} Is the overfitting problem affecting SNNs with various activation functions differently?

To answer the above mentioned questions, a number of experiments have been conducted to compare accuracy and loss function scores during training. The experiments have been conducted for 5 sparse levels and the dense network, for each of the seven Activation Functions (AFs): ReLU, Sigmoid, SELU, SReLU, Tanh, Softplus, Softsign. Sparsity levels used ranged from 71.2\% to 98.85\%.
    
This paper is divided into the following sections. The Background section explains the main concepts needed to understand the methods and findings. Next, Related Work section provides information on the researchers working on the topic of Sparse Neural Networks. Then, the methodology and results are discussed separately, providing details about a suggested framework for analysis of the differences in the impact of activation functions and the results of this implementation on the SET algorithm and CIFAR10 dataset, respectively. Lastly, the Conclusions and Future Work section summarizes the findings and describes suggested future research areas.

\section{Background}

\subsection{Artificial Neural Networks (ANN)}
Artificial Neural Networks is a family of networks built of input, hidden and output layers, inspired by the neural network of the biological brain. The most known example so far is the Multilayer Perceptron (MLP), a network consisting of densely connected layers of perceptrons \cite{ramchoun2016multilayer}. While there are more network models created and used nowadays, this research will mainly focus on the MLP.

\subsection{Sparse Neural Network (SNN)}
\begin{figure}[!hbt]
\includegraphics[width=0.42\textwidth]{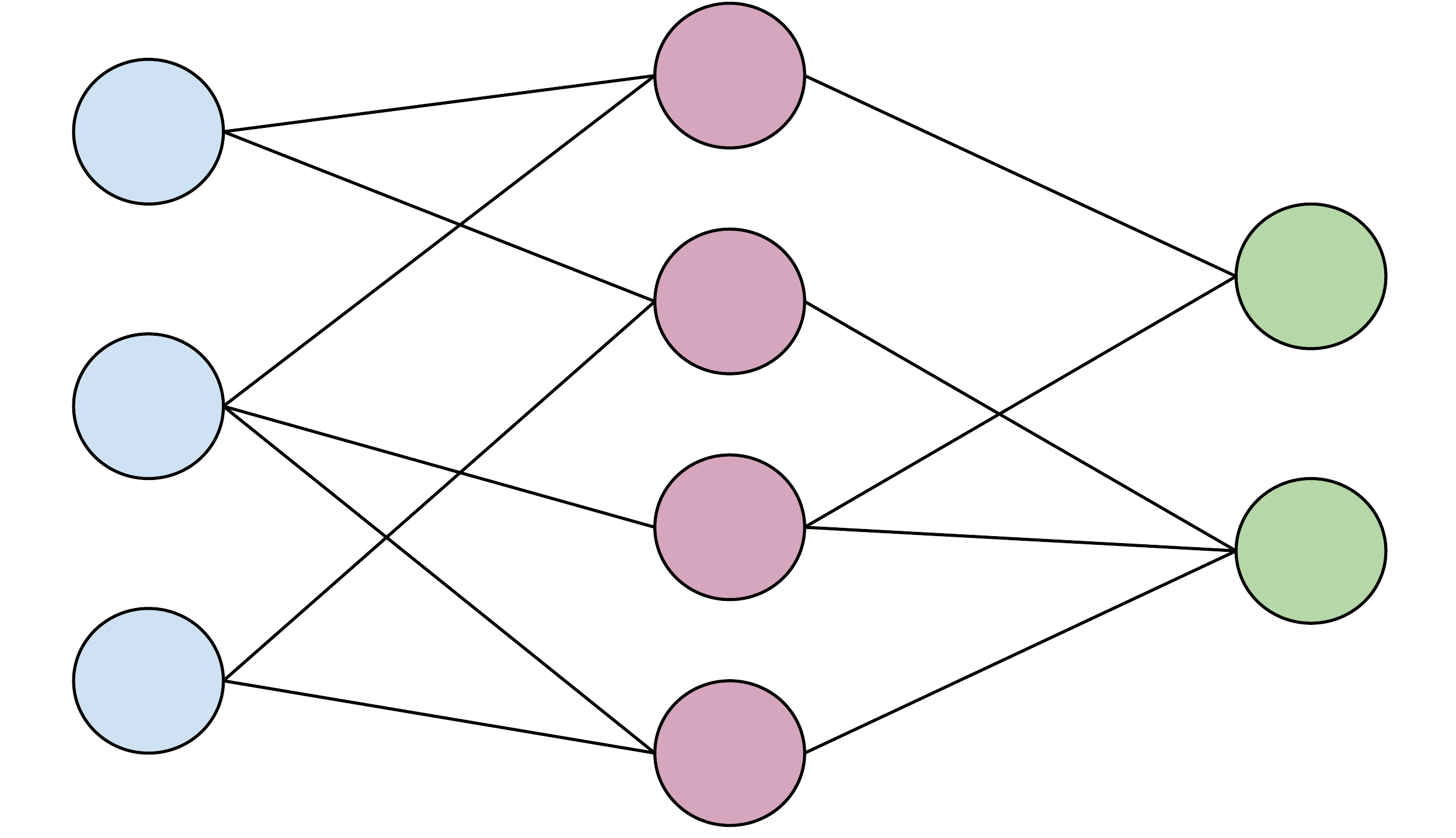} 
\caption{Sparsely connected ANN layers.}
\label{fig:sparse_ann}
\end{figure}

In comparison to Dense Neural Networks, SNNs drop some of the connections, thus limiting the computational complexity of the algorithm. That allows them to not only train faster but also enables bigger network architectures, which typically results in an accuracy increase. Figure \ref{fig:sparse_ann} visualizes an example of a sparsely connected architecture.

\subsection{Sparse Evolutionary Training (SET)}
Sparse Evolutionary Training is a method developed by Mocanu et al. \cite{mocanu2018scalable} allowing to build SNNs by first starting with sparsely connected layers, iteratively growing new weights in random locations and pruning the weights closest to 0. The framework suggested in this paper uses SET implementation as a code base for training and sparsity sweep evaluation.

\subsection{Activation Function (AF)}
\begin{figure}[!h]
    \includegraphics[width=0.46\textwidth]{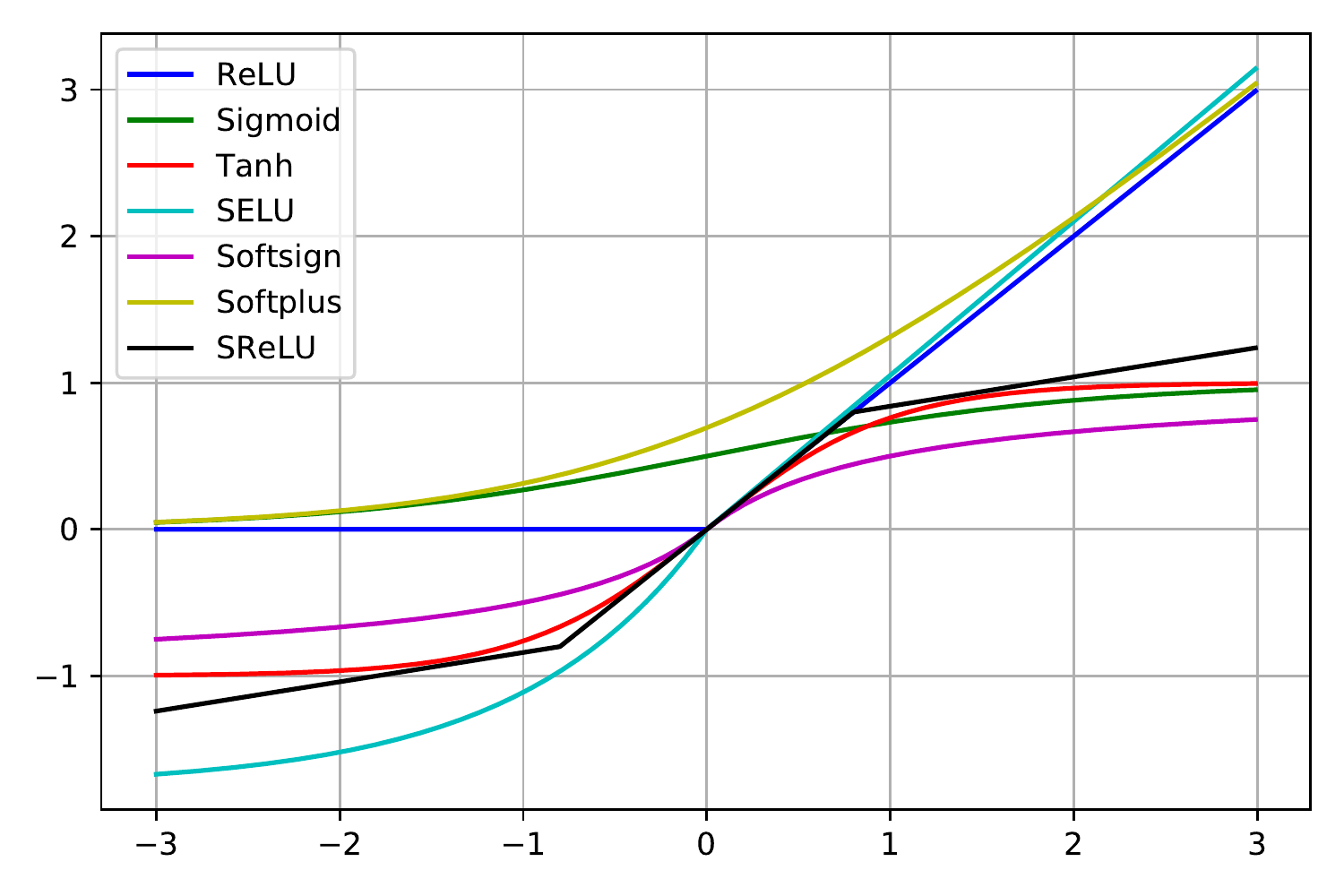}
    \caption{Plots of activation functions used.}
    \label{fig:af_plots}
\end{figure}

Activation Functions define the logic for each of the neurons within ANNs by modelling complex behaviour with non-linear mathematical functions \cite{sharma2017understanding}. Most popular functions are: Sigmoid , Hyperbolic Tangent (Tanh), Rectified Linear Unit (ReLU) \cite{ohn2019smooth}, Softsign and Softplus \cite{maksutov_2018} and a number of ReLU modifications like SELU and SReLU \cite{liu_2017}. Figure \ref{fig:af_plots} visualizes the functions.

\subsection{SReLU}
S-shaped rectified linear activation unit (SReLU) is a relatively little-known activation function suggested by Jin et al. \cite{jin2016deep} and used by Mocanu et al. \cite{mocanu2018scalable} in their implementation of the SET algorithm. It combines 3 linear functions defined by 4 dynamic parameters. Note that the SReLU plot in Figure \ref{fig:af_plots} is a sample presentation with set parameters. During training, the parameters are adjusted using back propagation \cite{jin2016deep}.

\subsection{Sparsity Sweep}
Sparsity Sweep is the relation between the accuracy and sparsity of the model. At a first thought, with higher sparsity of the model, some information is lost which implies a drop in accuracy. However, the performance of sparse neural networks can sometimes be greater than that of their dense counterparts with the same hyperparameters when they are trained with sparse training methods e.g. SET \cite{mocanu2018scalable}.

\newpage
\subsection{Overfitting}
\begin{figure}[!hbt]
\centering 
\includegraphics[width=0.4\textwidth]{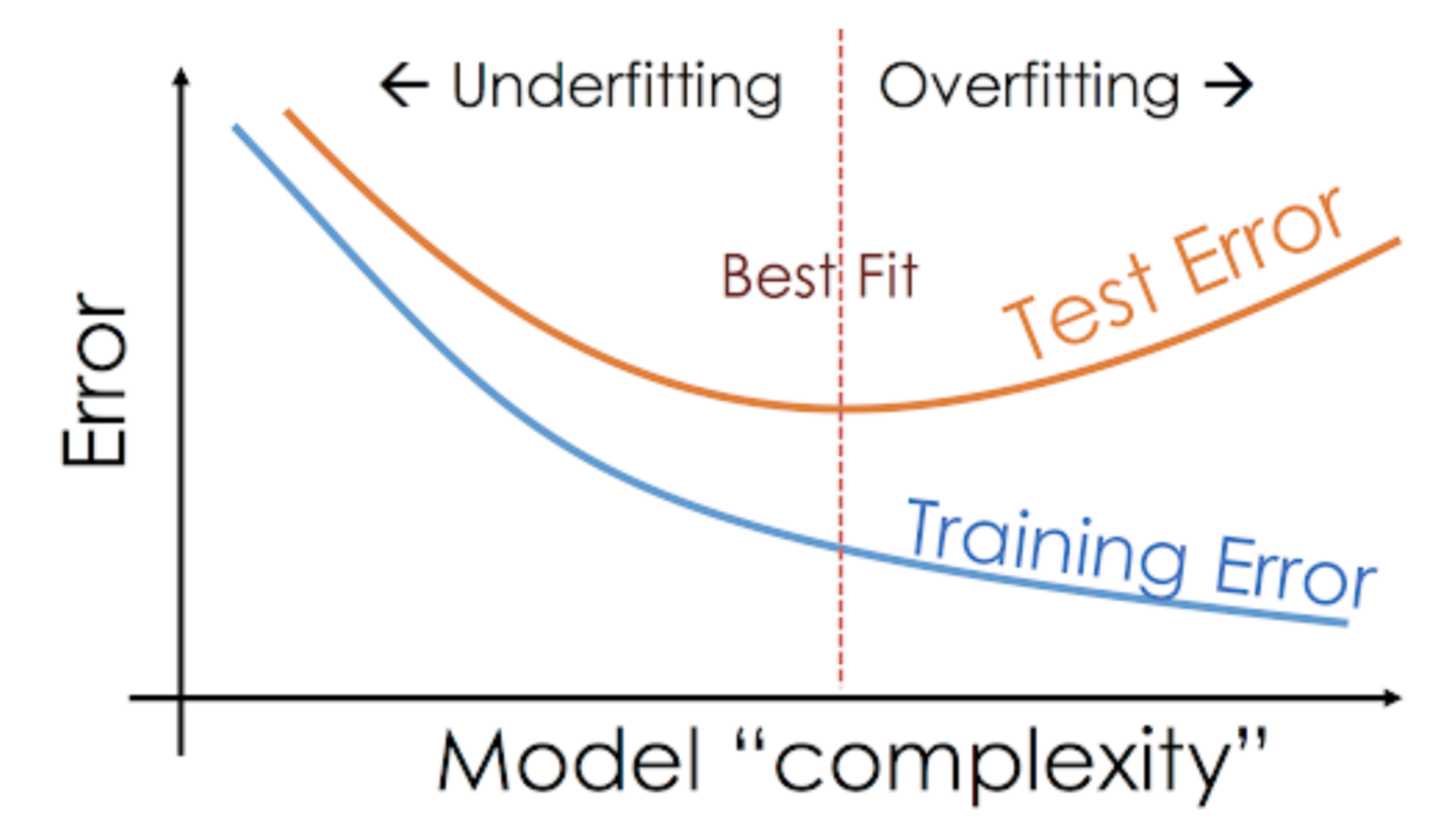} 
\caption{Overfitting example from \cite{overfitting}.}
\label{fig:overfitting}
\end{figure}

Over- and underfitting are two concepts in Machine Learning describing the discrepancy between training and validation (testing) set performance. Underfitting can be noticed when there is significant potential to improve the validation set accuracy, while overfitting means the effect when the algorithm tries to memorize the training data, while reducing the test accuracy. Figure \ref{fig:overfitting} explains the concepts in a visual form by showing example test and training errors during training.

\section{Related Work}
Researchers have conducted a significant amount of research on SNN implementations. To start with, in 1990, LeCun et al. \cite{lecun1990optimal} suggested pruning unimportant weights from Neural Networks to achieve better results and performance. Using those insights, in 2015, Han et al. \cite{han2015learning} suggested a training method which not only prunes unimportant weights but also retrains the model to tune the parameters. Experiments have shown \cite{han2015learning} that this allows for a much lower computational complexity, but a limitation of that method is the maximum size of a fully connected network that can be trained. To solve the issue, researchers have been working on sparse model training methods. In 2017, Mocanu et al. \cite{dcmocanuphd,mocanu2018scalable} have developed Sparse Evolutionary Training (SET) as a way to train a SNN from sparse instead of pruning a fully connected and trained ANN. The method removes weights close to 0 in a set interval and then distributes new ones randomly. Bellec et al. \cite{bellec2017deep} suggested a method for supervised rewiring of the network during training keeping the total number of connections low using stochastic sampling theory. The method has been called Deep Rewiring (Deep R) due to its extensive rewiring during the Deep Learning training process. Another approach, the Neural Network Synthesis Tool (NeST) initializes a sparsely connected network and updates parameters according to gradient information and performs pruning based on the magnitude of connections and neurons \cite{dai2019nest}. SET has proven to be efficient enough to run a network with over million neurons on a normal laptop \cite{liu2019sparse} and to be more accurate than Deep R \cite{dettmers2019sparse}. Lapshyna \cite{lapshyna2020sparse} proposed an improvement to the weight update process in SET by gradually removing connections during training related to the increase in the accuracy level. It was proved that the number of connections can be vastly limited with negligible loss of accuracy. Bird et al.\cite{bird2020dendritic} came up with another way to improve the current algorithms by using Dendritic Normalisation to improve learning performance, based on the SET algorithm. He proved that the costs and accuracy can be vastly improved, providing reliable quality for the future research. In 2019, Mostafa et al. \cite{mostafa2019parameter} suggested a method called Dynamic Sparse Reparametrization (DSR) presented as more efficient and accurate than the above-mentioned examples. It makes use of adaptive thresholds for the number of weights pruned and automatically reallocates parameters across layers, which avoids a fixed sparsity level. Sparse Momentum is a method redistributing weights relatively to mean momentum, both between layers and within each of the layers. DSR and Sparse Momentum have proven slightly more accurate than SET \cite{dettmers2019sparse}. Based on the works of their fellow researchers, a group of researchers from Google and DeepMind has come up with a Rigged Lottery (RigL) \cite{evci2019rigging} method for training SNNs with fixed complexity without accuracy loss. RigL removes connections based on weight magnitudes and adds new ones according to gradient information. This method has achieved the highest accuracy level and relatively low sparsity sweep, compared to other techniques.

While all the above-mentioned researchers have been working towards improvements in the accuracy of SNNs, they all have only tried improving the previous approaches by redistributing connections and associated weights. However, while there is quite some research on the impact of dense ANNs \cite{nwankpa2018activation},\cite{mishkin2015all},\cite{ramachandran2017searching},\cite{glorot2010understanding},\cite{jin2016deep}, little or no research has been conducted on the impact of the activation function choice on SNNs' performance, most probably because this research direction is still new and under development. 
The approach suggested in this paper is mainly based on SET but can be also applied to other works, which could provide insightful overview on the differences between when many activation functions are taken into account.

\section{Evaluation framework}
In this research, a framework for comparing performance and the sparsity effect between activation functions is proposed. Although the explained example is based on the SET algorithm implementation \cite{mocanu2018scalable}, we believe that, with some adjustments, it can also be used for other sparse training methods.\\
First of all, all experiments used the same hyperparameters stated in Table \ref{tab:hyperparam}, ensuring the performance changes depend solely on the adjusted parameters - epsilon and the activation function. Although they can be adjusted, one should make sure to keep them consistent across all experiments, as they might impact the outcomes significantly.

\begin{table}[hb!] 

\begin{tabular}{@{}lll@{}}
\toprule
\begin{tabular}[c]{@{}l@{}}Hyperparameter\end{tabular}
& Value                    \\ \midrule
Learning rate       & 0.01      &           \\
Optimiser           & SGD       &           \\
Momentum            & 0.9       &           \\
$\upzeta$ (zeta)    & 0.3       &           \\
Batch size          & 100       &           \\

\begin{tabular}[c]{@{}l@{}}Loss Function\end{tabular}
& 
\begin{tabular}[c]{@{}l@{}}Categorical Cross-entropy\end{tabular} \\ 
Dropout rate        & 0.3                        \\
\hline
\end{tabular}
\caption{Hyperparameters used in training of MLP \cite{goodfellow2016}.}
\label{tab:hyperparam}
\end{table}

Moreover, the SET algorithm implementation needs to be adjusted so that the accuracy and loss function values are saved for both the training and validation sets after each of the training epochs.
Then, models are trained for all combinations of selected activation functions and epsilon values corresponding to the desired sparsity levels.
Sparsity means simply the probability of a connection between neurons to be removed and thus, it is the opposite of density. Note that this definition is closely related to SET since it removes weights randomly and thus other training techniques may implement sparsity differently. 
\begin{equation}
    \label{eq:sparsity}
    Sparsity = 1 - density
\end{equation}
Since sparsity is dependent both on the epsilon value and the architecture of the algorithm , we can derive the equation for epsilon $\upepsilon$ from equation (1) of SET \cite{mocanu2018scalable} as
\begin{equation}
\label{eq:epsilon}
\upepsilon = (1 - sparsity) \times (\dfrac{ n_{l} \times n_{l+1}}{n_{l} + n_{l+1}})
\end{equation}

where  $n_{l}$ and $n_{l+1}$ mean the dimensions of consecutive layers. This allows us to easily calculate epsilon for all needed sparsity levels. Table \ref{tab:nparameters} shows number of incoming connections per layer, depending on the sparsity of the network. Algorithm \ref{alg:pseudo_framework} describes steps needed to optimize the evaluation procedure. Saving results in clearly declared directories allows for automatic chart generation for all activation functions and sparsity levels.

\begin{table}[!hbt]
\centering
\begin{tabular}{@{}l|l|lll@{}}
\toprule
\multirow{2}{*}{Sparsity} & \multirow{2}{*}{Epsilon} & \multicolumn{3}{l}{Parameters {[}\#{]}} \\ \cmidrule(l){3-5} 
 &  & 1st layer & 2nd layer & 3rd layer \\ \midrule
98.85\% & 10 & 141312 & 46000 & 46000 \\
97.12\% & 20 & 353895 & 115200 & 115200 \\
94.25\% & 50 & 706560 & 230000 & 230000 \\
88.50\% & 100 & 1413120 & 460000 & 460000 \\
71.20\% & 500 & 3538944 & 1152000 & 1152000 \\
dense & n.d. & 12288000 & 4000000 & 4000000 \\ \bottomrule
\end{tabular}
\caption{Number of parameters between layers per sparsity}
\label{tab:nparameters}
\end{table}

\begin{algorithm}
  \centering
  \caption{Proposed Sparsity Sweep evaluation framework}
  \label{alg:pseudo_framework}
  \begin{algorithmic}[0]
	\State set hyperparameters
	\State adjust parameters to be saved
    \For{each activation function (AF)}
        \State set activation function 
    	\State set output location for results
    	\For{each sparsity level}
    	    \State set $\upepsilon$ value using equation \ref{eq:epsilon}
    	    \State train model
            \EndFor
        \EndFor
    \State produce performance charts
    \State produce sparsity sweep charts
    \State produce overfitting charts
	\end{algorithmic}
\end{algorithm}

\begin{figure*}[!hbt]
    \centering
    \begin{subfigure}{.495\textwidth}
        \includegraphics[width=\textwidth]{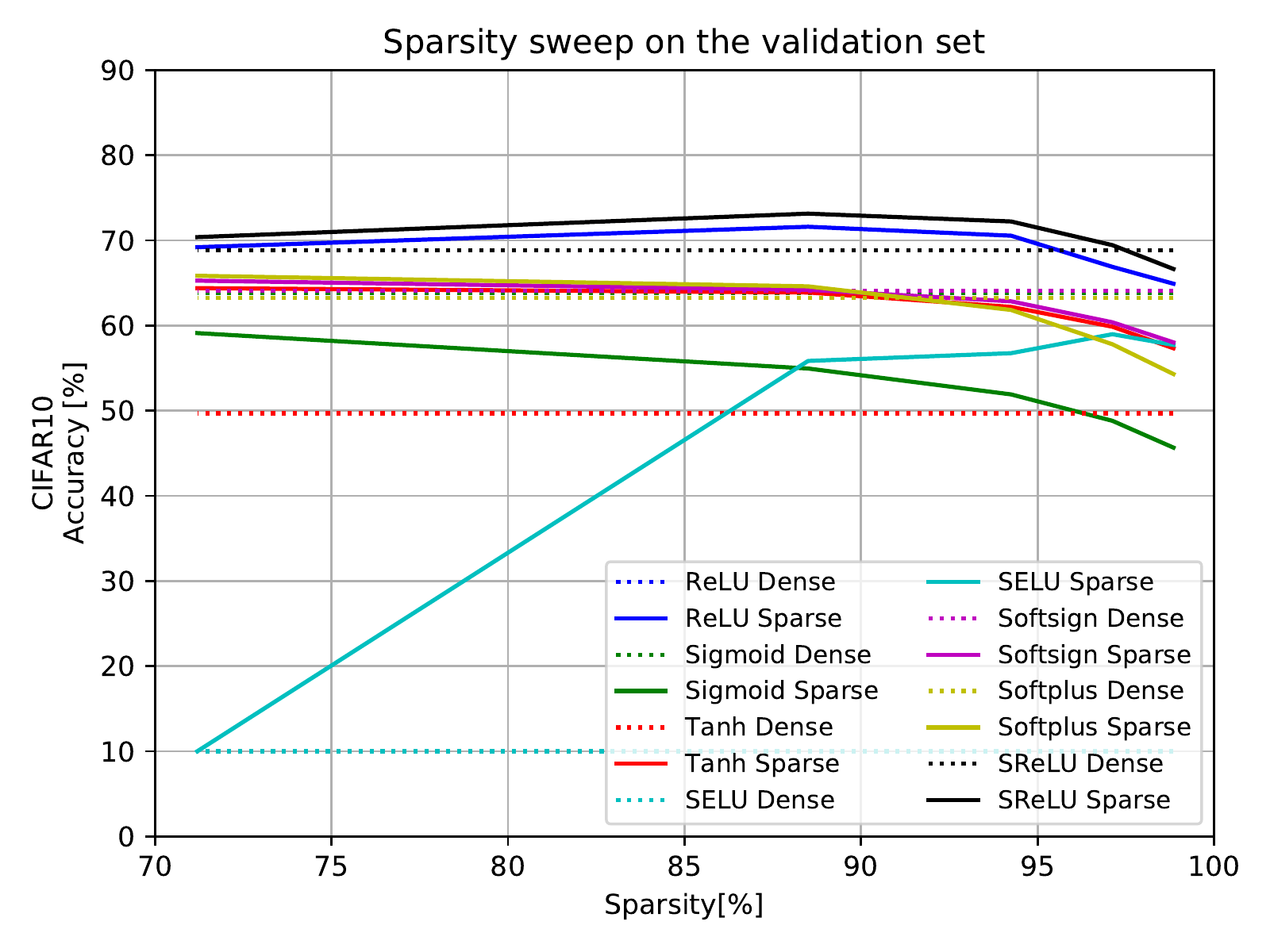}
        \caption{Validation set}
        \label{subfig:sparsity_sweep_validation}
    \end{subfigure}
    \begin{subfigure}{.495\textwidth}
        \includegraphics[width=\textwidth]{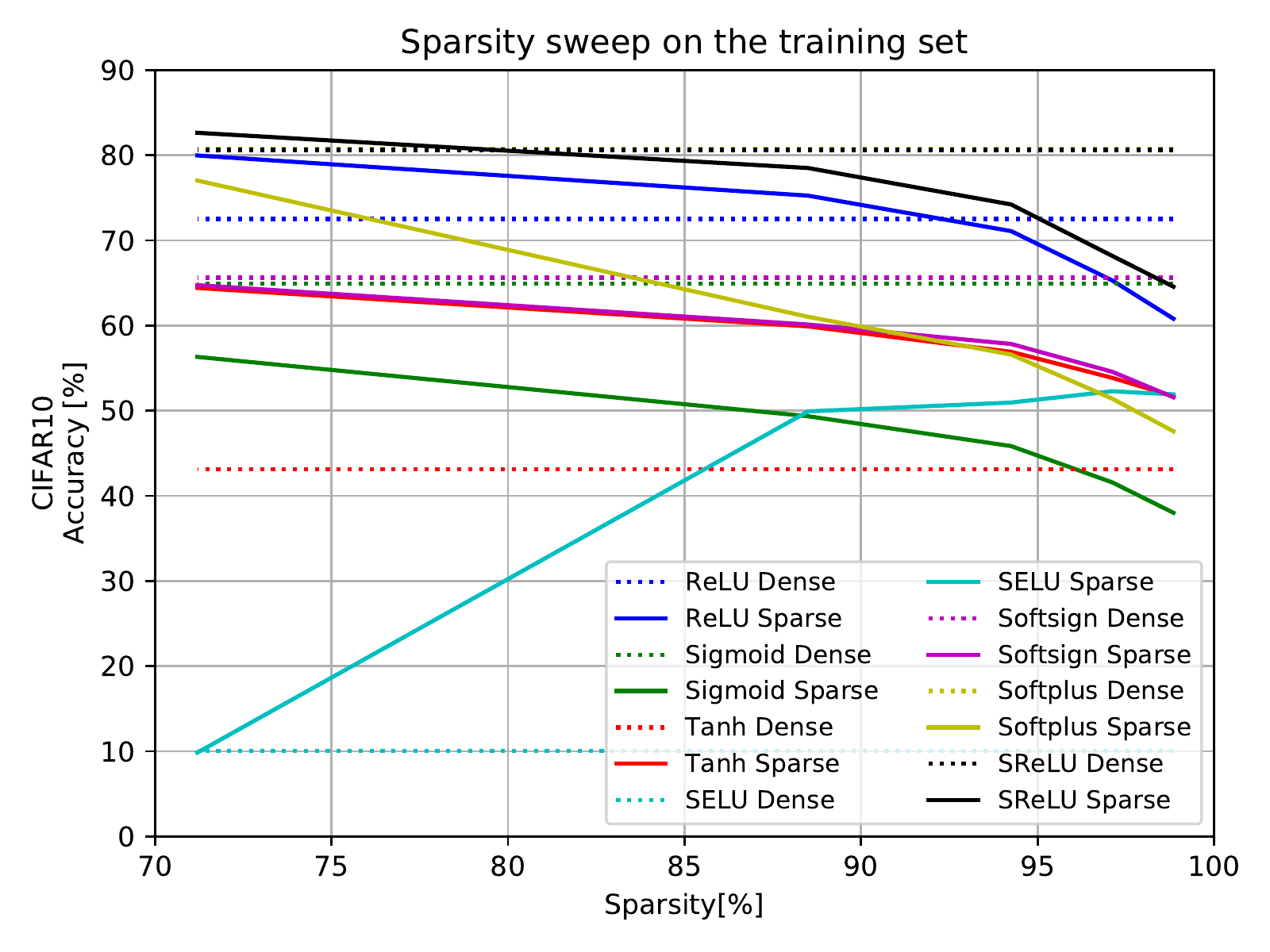}
        \caption{Training set}
        \label{subfig:sparsity_sweep_training}
    \end{subfigure}
    \caption{Accuracy sparsity sweep}
    \label{fig:sparsity_acc}
\end{figure*}

\subsection{Overfitting function}
Overfitting occurs when a network is trained in too much detail, trying to fit the noise in the data into their model. Most of the time, scientists try to analyze this effect by comparing training and validation set accuracy scores over the training period. However, since several activation functions are compared in this study, a new function displaying overfitting as the difference between the scores is introduced:
\begin{equation}
    o = acc_{t} - acc_{v}
    \label{eq:overfitting}
\end{equation}
where $o$, $acc_{t}$ and $acc_{v}$ represent overfitting, training set accuracy and validation set accuracy, respectively. Overfitting values (Eq. \ref{eq:overfitting}) above 0 suggest that the network is overfitting, while negative values might mean the opposite - underfitting. 
Please note that in the beginning of training, the network can show significantly unstable results and thus we should focus on the long-term trend of the function rather than its details. 

\section{Results}
To gain deeper insights, researchers often need to take a look at their results from many perspectives. Therefore, a number of plots have been prepared and analysed but in order to convey the most important message, only a handful is shown below. A curious reader can find other charts in the appendix. In this section, the implementation details as well as the outcomes of the study will be discussed.

\newpage
\subsection{Implementation and architecture}
Within this research, the framework suggested above has been implemented using the adjusted SET-MLP-Keras-Weights-Mask project developed as a part of and presented in \cite{mocanu2018scalable}. Separate scripts included in the project have been used to train the sparse and dense networks.
The project uses Keras for model training and Python packages NumPy and Matplotlib to visualize the results. It is important to note that while the results observed with this implementation resemble the behavior of truly sparse networks, at the moment (June, 2020) Keras does not support SNNs and thus the SET implementation used in this research masks weights to 0 to achieve the same structure. The experiments have been conducted on the CIFAR10 dataset, which consists of 60000 images split into 10 classes, 6000 images each. As the dimensions of the images are 32x32[px] and represented in 3 channel colors, the input layer of the model has $n_{0}$ number of input parameters as follows:
\begin{equation}
    n_{0} = 32\times32\times3 = 3072 
\end{equation}
The hidden layers have been set in the research done by Mocanu et al. \cite{mocanu2018scalable} to 3 layers, 4000, 1000 and 4000 neurons each, respectively. Lastly, the output layer contains 10 neurons, corresponding to 10 classes in the CIFAR10 dataset
. All models were trained for 500 epochs and used the same hyperparameters but the epsilon value and activation function.


\subsection{Sparsity Sweep}

For each activation function, a model has been trained using 6 different sparsity levels, from 98.85\% (only 1.15\% connections left) to the dense level. Scores for the accuracy as well as the loss function value were collected after each training epoch for both training and validation sets. Table \ref{tab:accuracy} shows the final accuracy results for each of the models after 500 epochs which are visualized in the Figure \ref{subfig:sparsity_sweep_validation}. It is clear to note that SReLU has performed best overall, while ReLU performed almost as well, with only a slight difference in accuracy on all sparsity levels. Networks using Softsign, Tanh and Softplus scored slightly worse and more importantly, their performance decreased with increased sparsity, meaning their optimal sparsity is much lower than for ReLU and SReLU. The case of SELU is significantly standing out because at the dense levels, the network could not predict anything at more than the random chance (10\%) after 500 epochs of training, while very sparse networks trained with SELU performed better than those with Sigmoid or Softplus and just as well as Tanh and Softsign. 

\begin{table}[!hbt]
\centering
\resizebox{\columnwidth}{!}{%
\begin{tabular}{l|llllll}
\hline
\multicolumn{1}{c|}{\multirow{2}{*}{Activation}} & \multicolumn{6}{l}{Accuracy {[}\%{]} on the validation set at sparsity level:} \\
\multicolumn{1}{c|}{} & 98.85\% & 97.12\% & 94.25\% & 88.50\% & 71.20\% & dense \\ \hline
ReLU & 64.91 & 66.89 & 70.55 & \textbf{71.6} & 69.2 & 63.21 \\
Sigmoid & 45.66 & 48.81 & 51.91 & 54.95 & 59.11 & \textbf{63.8} \\
Tanh & 57.33 & 59.86 & 62.19 & 63.88 & \textbf{64.39} & 49.69 \\
Softplus & 54.3 & 57.8 & 61.83 & 64.6 & \textbf{65.84} & 63.27 \\
Softsign & 58.02 & 60.38 & 62.84 & 64.19 & \textbf{65.27} & 64.08 \\
SELU & 57.72 & \textbf{58.98} & 56.75 & 55.84 & 10 & 10 \\
SReLU & 66.64 & 69.44 & 72.22 & \textbf{73.14} & 70.38 & 68.86
\end{tabular}
}
\caption{Accuracy [\%] on the validation set after 500 epochs}
\label{tab:accuracy}
\end{table}

\begin{figure*}[!hbt]
    \centering
    \begin{subfigure}{.46\textwidth}
        \centering
        \includegraphics[width=\textwidth]{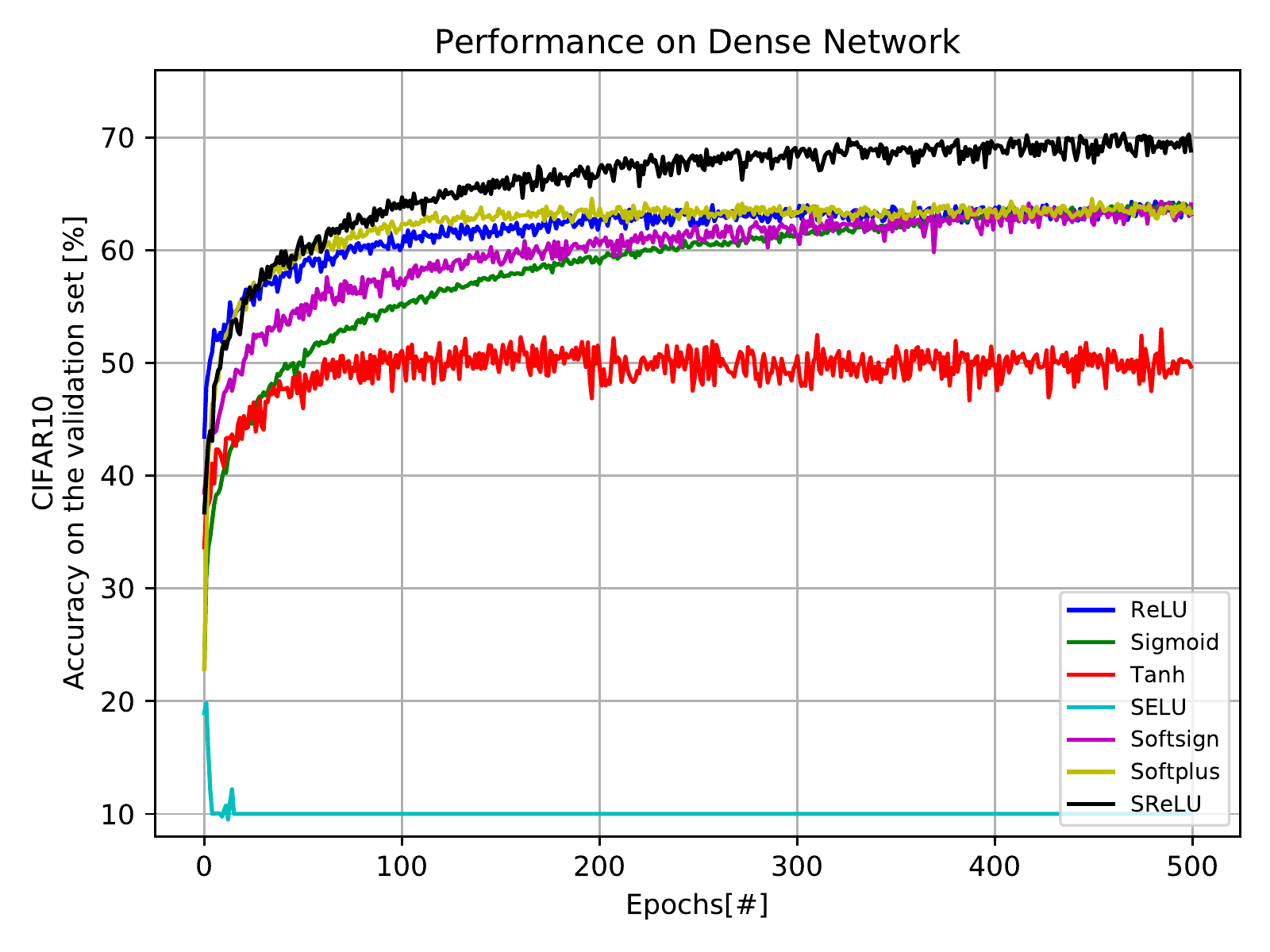}
        \caption{Dense}
        \label{subfig:performance_dense}
    \end{subfigure}
    \quad
    \begin{subfigure}{.46\textwidth}
        \centering
        \includegraphics[width=\textwidth]{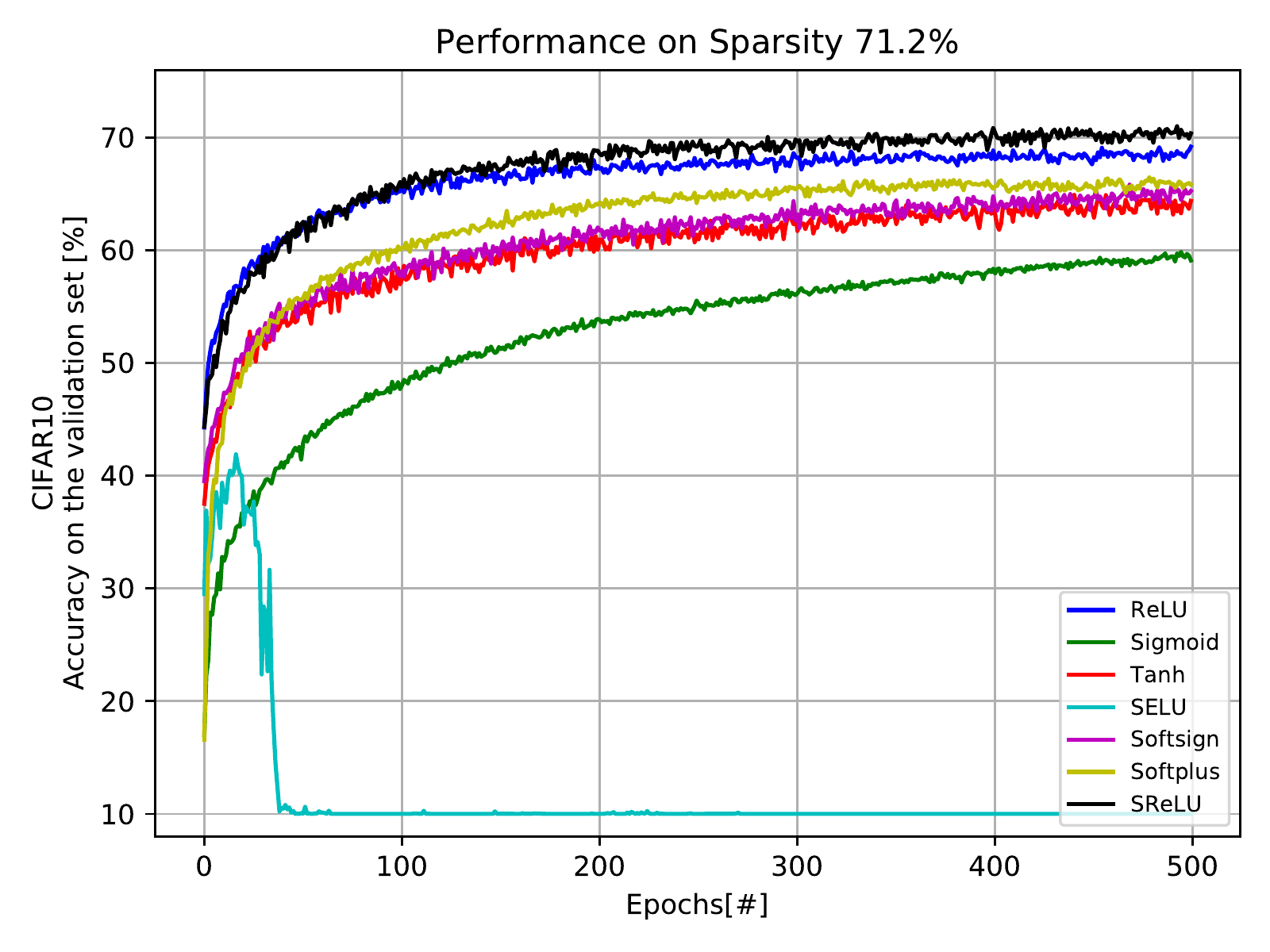}
        \caption{Sparsity 71.2\%}
        \label{subfig:performance_e500}
    \end{subfigure}
    \vskip\baselineskip
    \begin{subfigure}{.46\textwidth}
        \centering
        \includegraphics[width=\textwidth]{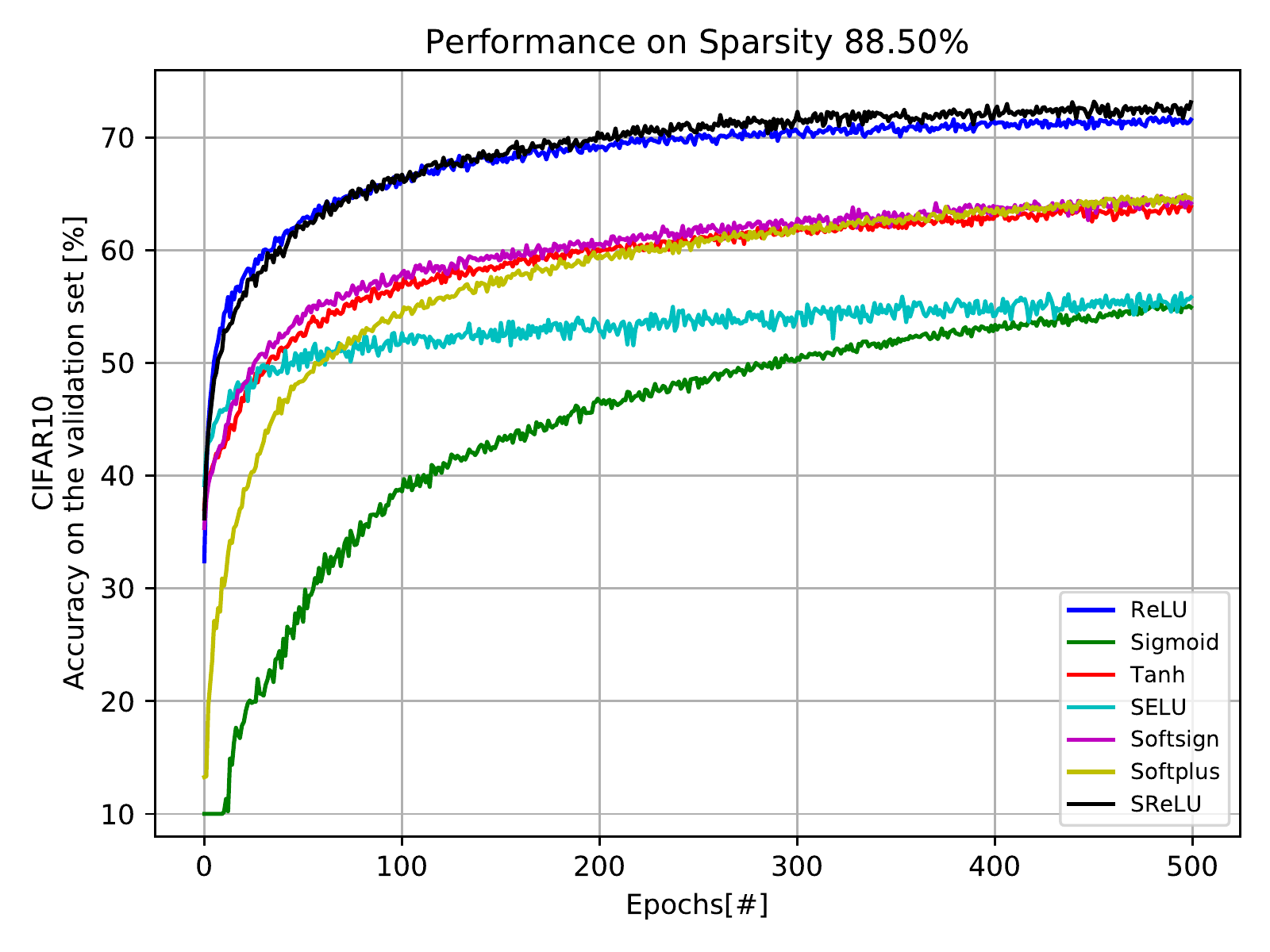}
        \caption{Sparsity 88.5\%}
        \label{subfig:performance_e100}
    \end{subfigure}
    \quad
    \begin{subfigure}{.46\textwidth}
        \centering
        \includegraphics[width=\textwidth]{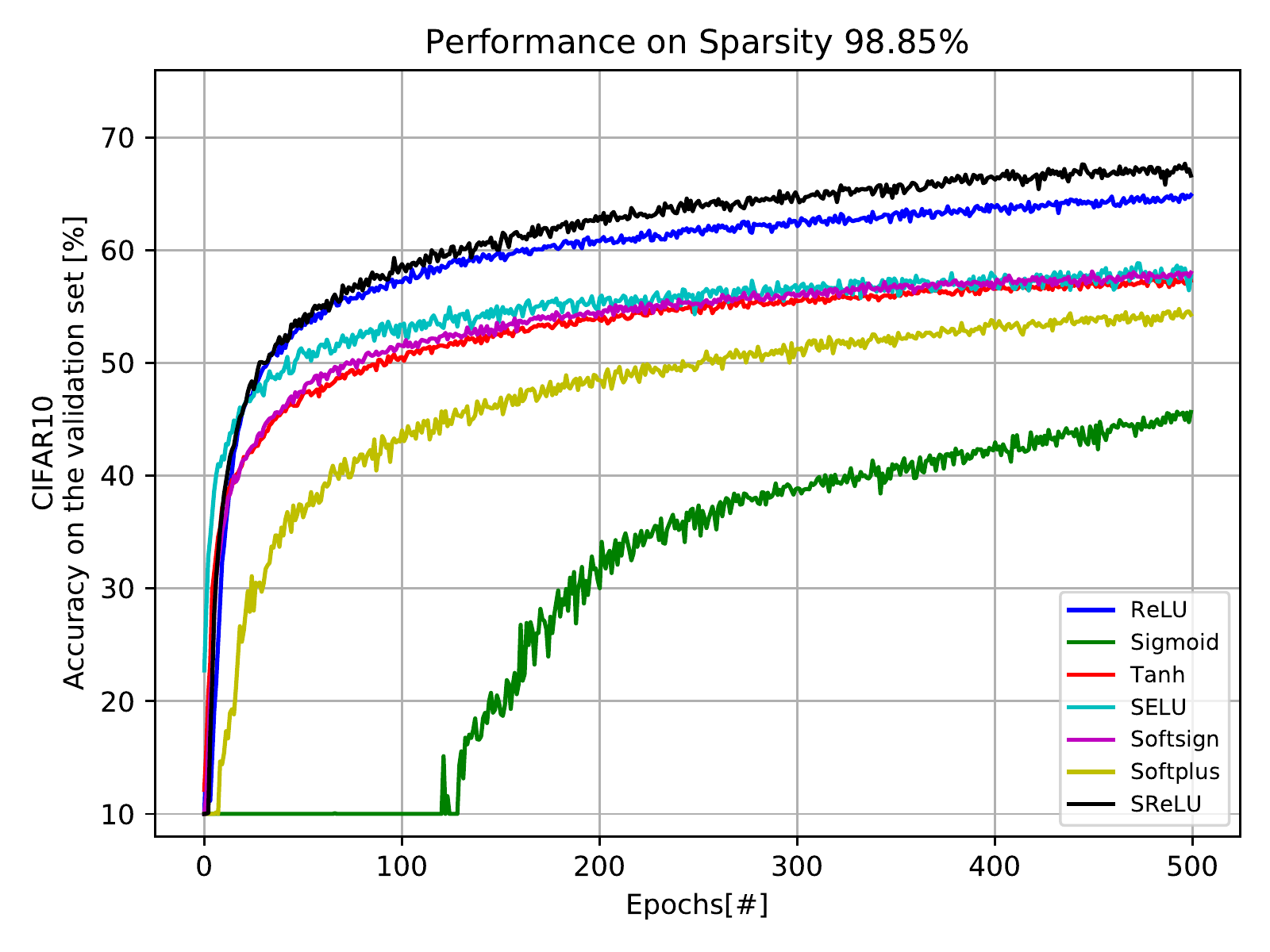}
        \caption{Sparsity 98.85\%}
        \label{subfig:performance_e10}
    \end{subfigure}
    \caption{Comparison of models' performance.}
    \label{fig:performance_comparison}
\end{figure*}

\subsection{General Performance}
When it comes to the training performance, it is best to look at the performance charts. Figure \ref{fig:performance_comparison} shows four charts which point out the most notable differences between the activation functions across the four selected sparsity levels.

Clearly, SReLU is the winner when it comes to accuracy, but it does not gain much better performance at optimal sparsity levels. ReLU, on the other hand, gains a lot of accuracy when trained on a sparse network, while it performs only as well as Softplus, Softsign, SELU and Sigmoid in the dense setting. Interestingly, Sigmoid is an example of a function that performs fine on dense network, but did not work well with sparse networks - its learning progress was somewhat delayed proportionally to the sparsity level. SELU is an example of a network that shows the opposite effect - while it performs fairly well on sparse networks, the training dies out after several training epochs. The behavior will be further discussed in the discussion section. When it comes to extremely sparse networks e.g. Sparsity 98.85\% at Figure \ref{subfig:performance_e10}, we can notice that while general accuracy dropped noticeably, the difference in accuracy varies significantly, depending on the activation function. Most noticeable drop in accuracy by the sparsity affected the models using Sigmoid and Softplus, while the model using SELU activation function scored higher than its dense counterparts.

To better understand the reasons behind the differences in performance of the examined activation functions, one needs to look at the problem from many angles. Plots visualizing the data from the function-specific point of view can be found in the appendix section \ref{sec:performance}.


\newpage
\subsubsection{SELU}
\begin{figure}[!hb]
  \centering
  \includegraphics[width=.445\textwidth]{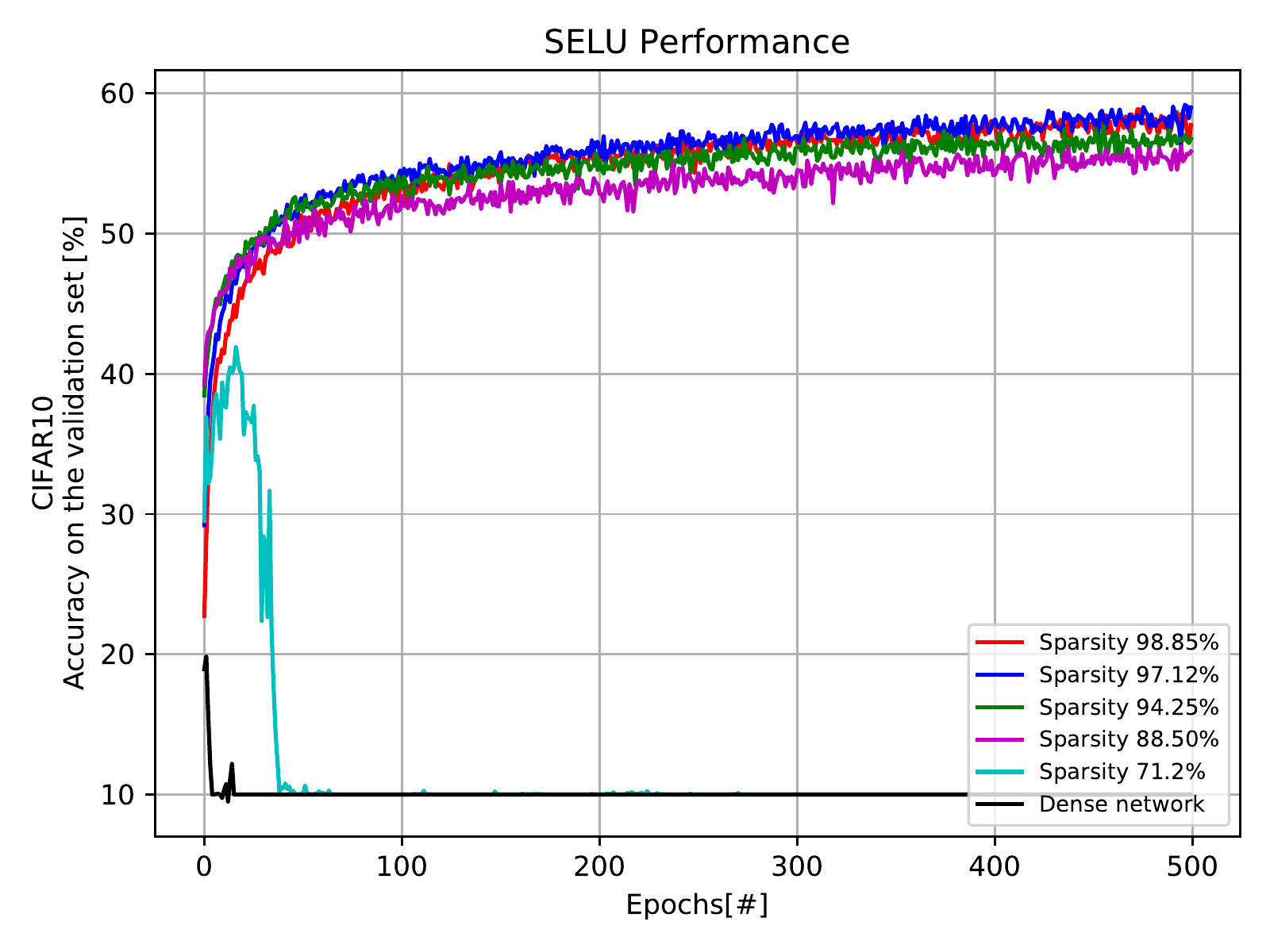}
  \caption{SELU accuracy on the validation set}
  \label{fig:performance_selu}
\end{figure}

Unexpected behavior of SELU is clearly noticeable on the dense levels. The performance chart of SELU in Figure \ref{fig:performance_selu}, clearly shows that the dense- and the sparsest network (Sparsity 71.2\%) experience an accuracy drop to the random level (10\%) after a couple of training epochs. Our hypothesis is that the learning rate used in the training was too high. As definite conclusions cannot be gained from just the networks trained in this study, a study on many dense levels is needed to find an explanation of the issue.

\begin{figure*}[!hbt]
\centering
\begin{subfigure}{.48\textwidth}
  \centering
  \includegraphics[width=\linewidth]{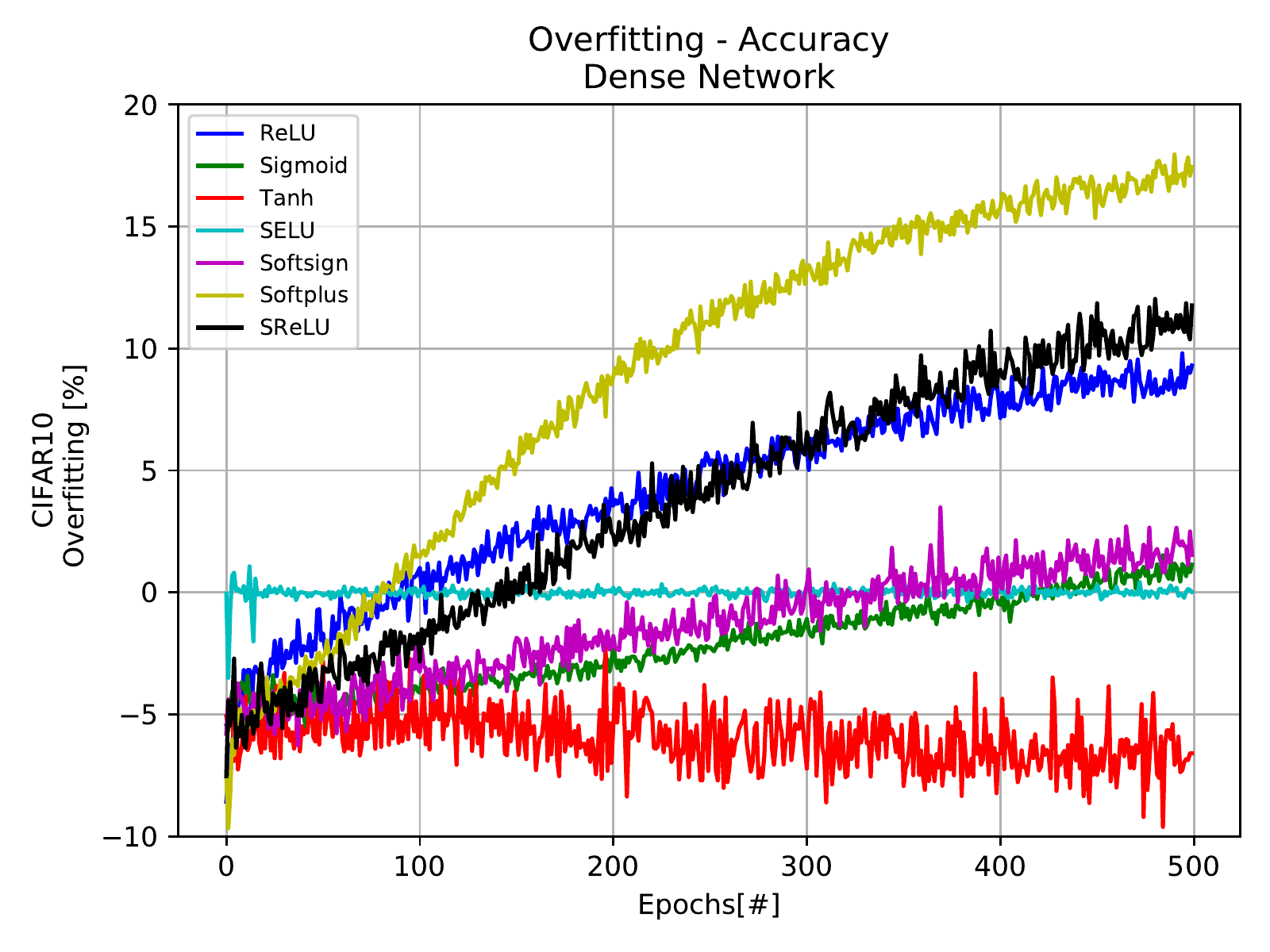}
  \caption{Dense}
  \label{subfig:overfitting_dense}
\end{subfigure}%
\quad
\begin{subfigure}{.48\textwidth}
  \centering
  \includegraphics[width=\linewidth]{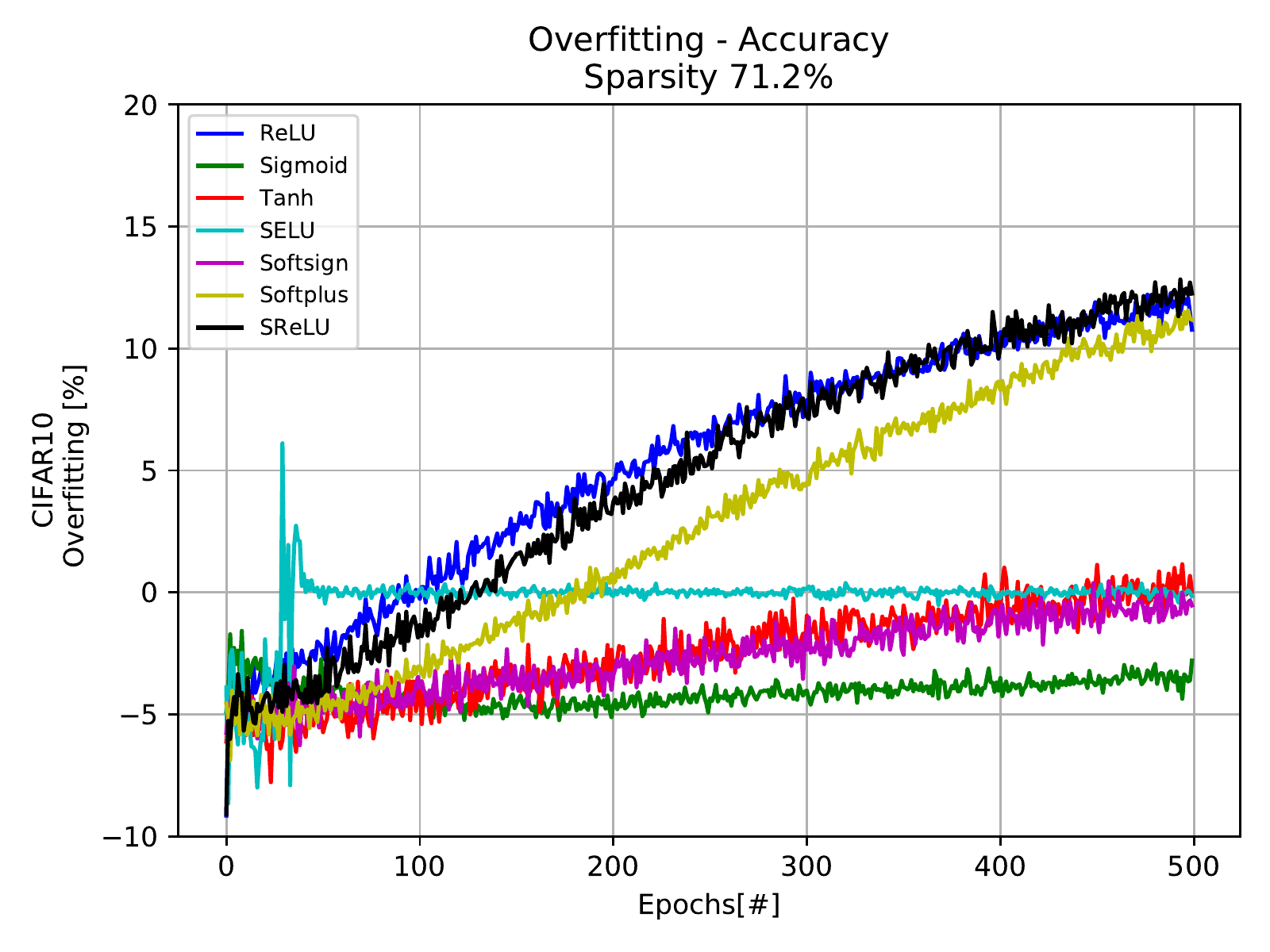}
  \caption{Sparsity 71.2\%}
  \label{subfig:overfitting_e500}
\end{subfigure}%

\vskip\baselineskip

\begin{subfigure}{.48\textwidth}
  \centering
  \includegraphics[width=\linewidth]{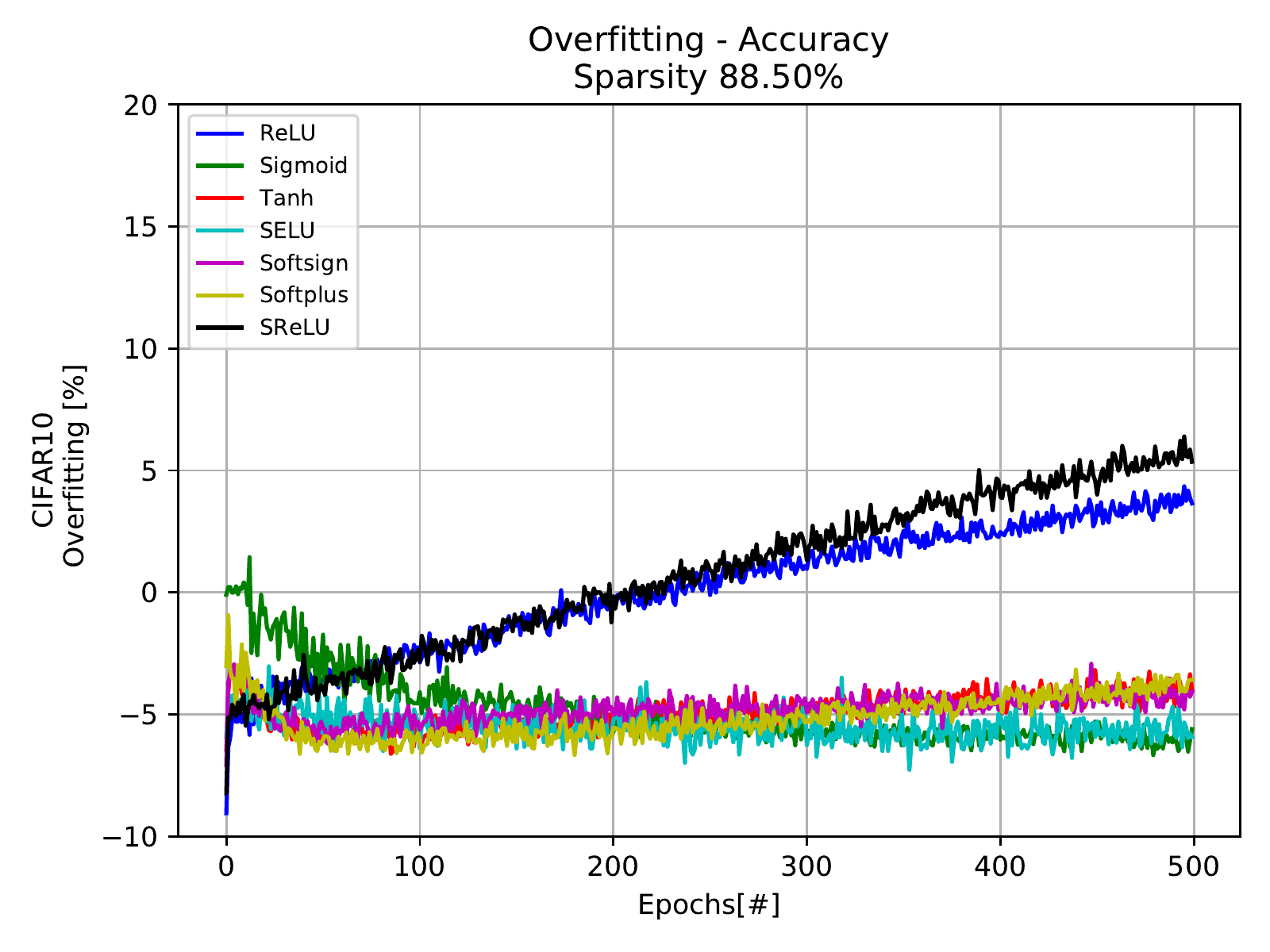}
  \caption{Sparsity 88.5\%}
  \label{subfig:overfitting_e100}
\end{subfigure}%
\quad
\begin{subfigure}{.48\textwidth}
  \centering
  \includegraphics[width=\linewidth]{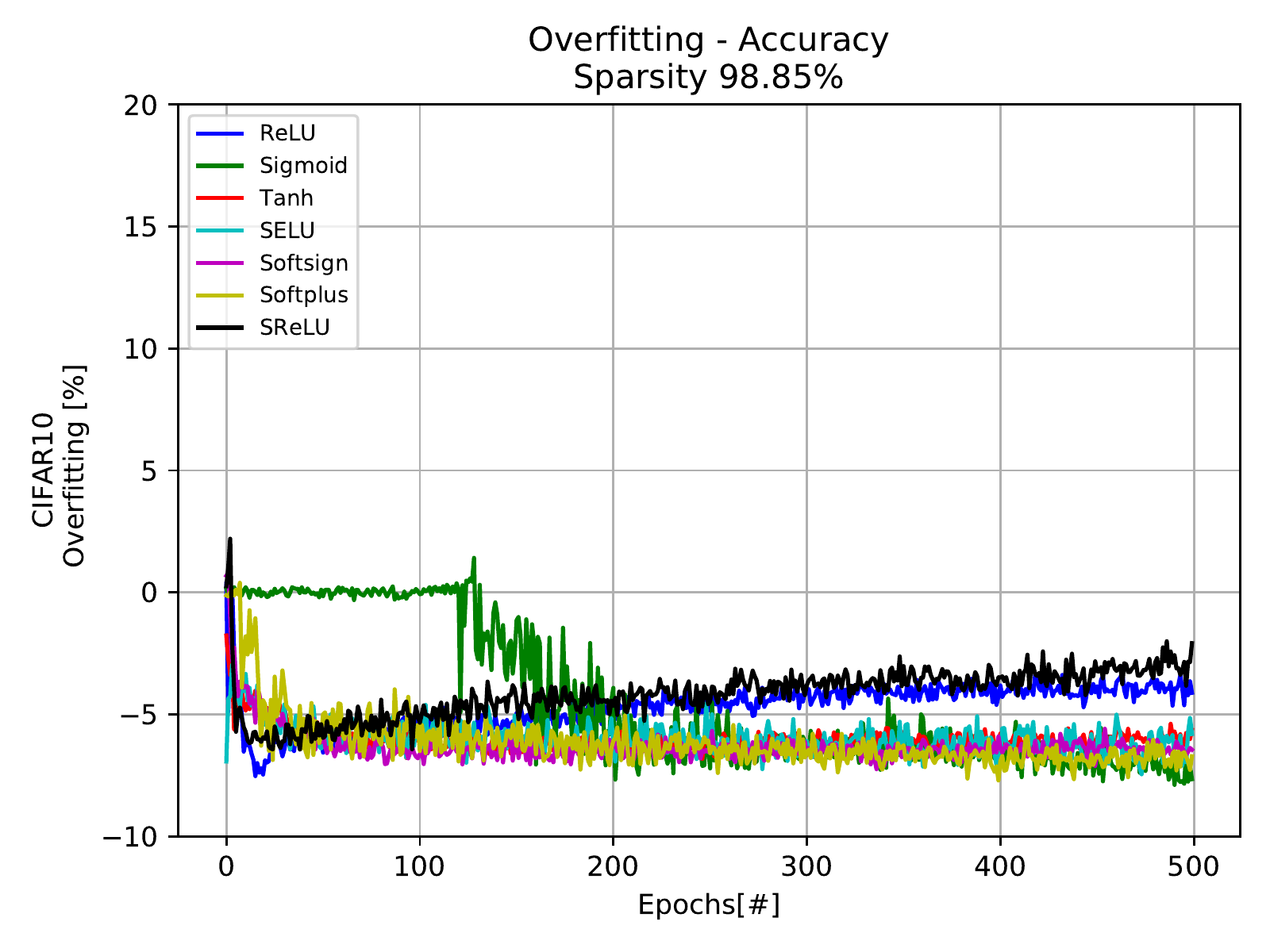}
  \caption{Sparsity 98.85\%}
  \label{subfig:overfitting_acc_e10}
\end{subfigure}%

\caption{Overfitting effect per sparsity level.}
\label{fig:overfitting_acc}
\end{figure*}

\newpage
\subsection{Overfitting}

Charts in Figure \ref{fig:overfitting_acc} show the above-mentioned overfitting function results for four selected sparsity levels. We can notice that for the dense level (Fig. \ref{subfig:overfitting_dense}), the difference between functions is significant. Tanh is underfitting, Softplus and Sigmoid fit relatively well, while ReLU, SReLU and Softplus are significantly overfitting. The overfitting effect for SELU is not analysed on dense levels due to the network training issue mentioned before.\\
Since overfitting is caused by over-trained models, it is reasonable to assume that with increased sparsity the overfitting effect will decrease, which is indeed the case. However, not all functions react in the same way. At sparsity level 71.2\% (Fig. \ref{subfig:overfitting_e500}), most functions show smaller overfitting effect while the Hyperbolic Tangent (Tanh) seems to achieve a good fit. We can notice that Sigmoid started underfitting, while SReLU and ReLU plots seem to show barely any difference when compared to the dense plots. At sparsity 88.5\% (Fig. \ref{subfig:overfitting_e100}), the overfitting effect is even less noticeable - most functions are underfitting, while only ReLU and SReLU are slightly overfitting. In case of the extremely sparse networks with sparsity 98.85\% (Fig. \ref{subfig:overfitting_acc_e10}), all functions were underfitting.

\newpage
\section{Conclusions and Future Work}
The framework suggested allows for comparisons between activation functions and their effects on the sparsity sweep of SNNs, while the experiments conducted provide insights into the impact of the selected activation functions on the accuracy of models trained with SET. Clearly, SReLU, which was used in Mocanu's implementation of SET \cite{mocanu2018scalable}, performs best at all sparsity levels, but the sparsity does not affect its achieved accuracy significantly. Other functions like ReLU, Tanh, Softplus achieve noticeably higher performance at sparsity levels 71.2\% and 88.5\%, higher than at the dense levels. While in general, most activation functions perform best at sparsity level 71.2\%, ReLU and SReLU performed best at sparsity 88.5\% and SELU achieved best performance at the extremely high sparsity level 98.85\%. Sigmoid is an exception there, as its performance decreases for sparser networks. Our hypothesis is that it is caused by Sigmoid not being zero-centered. When it comes to under- and overfitting, at higher sparsity levels, most functions start underfitting but they are not affected by the sparsity equally. SReLU and ReLU seem to have highest optimal point of sparsity when they achieve the best training fit, while the overfitting effect of Softplus was most affected by the increased sparsity.\\
Hopefully, researchers will find this paper useful to get insight and inspirations on the impact of activation functions on the sparsity sweep of SNNs. Naturally, this research does not cover all the questions around this topic and therefore, more research in the area would help answer the questions. \\In the future, researchers could work on the following issues:
\begin{enumerate}
    \item The experiments shall be repeated on datasets other than CIFAR10 to compare the results.
    \item More experiments could be done on more dense levels around 50-90\%. This would provide more accurate insights for optimal sparsity levels of each of the activation functions.
    \item The training issue of SELU needs to be examined with a deeper mathematical analysis on a number of dense levels.
    \item More work can be done to better understand the underlying reasons for the differences between the effects of those activation functions.
    \item Naturally, other AFs should still be tested to better understand the similarities between groups of AFs
\end{enumerate}

\section{Acknowledgements}
I would like to express great appreciation to Decebal Mocanu for his guidance and useful directions which allowed me to focus on what was most important. Additionally, I would like to thank Selima Curci, who helped me overcome technical obstacles with implementation of the SReLU activation function.

\clearpage
\balance
\bibliographystyle{abbrv}
\bibliography{ADubowski_thesis} 

\begin{thebibliography}{10}

\bibitem{bellec2017deep}
G.~Bellec, D.~Kappel, W.~Maass, and R.~Legenstein.
\newblock Deep rewiring: Training very sparse deep networks.
\newblock {\em arXiv preprint arXiv:1711.05136}, 2017.

\bibitem{bird2020dendritic}
A.~D. Bird and H.~Cuntz.
\newblock Dendritic normalisation improves learning in sparsely connected
  artificial neural networks.
\newblock {\em bioRxiv}, 2020.

\bibitem{dai2019nest}
X.~Dai, H.~Yin, and N.~K. Jha.
\newblock Nest: A neural network synthesis tool based on a grow-and-prune
  paradigm.
\newblock {\em IEEE Transactions on Computers}, 68(10):1487--1497, 2019.

\bibitem{denil2013predicting}
M.~Denil, B.~Shakibi, L.~Dinh, M.~Ranzato, and N.~De~Freitas.
\newblock Predicting parameters in deep learning.
\newblock In {\em Advances in neural information processing systems}, pages
  2148--2156, 2013.

\bibitem{dettmers2019sparse}
T.~Dettmers and L.~Zettlemoyer.
\newblock Sparse networks from scratch: Faster training without losing
  performance.
\newblock {\em arXiv preprint arXiv:1907.04840}, 2019.

\bibitem{evci2019rigging}
U.~Evci, T.~Gale, J.~Menick, P.~S. Castro, and E.~Elsen.
\newblock Rigging the lottery: Making all tickets winners.
\newblock {\em arXiv preprint arXiv:1911.11134}, 2019.

\bibitem{glorot2010understanding}
X.~Glorot and Y.~Bengio.
\newblock Understanding the difficulty of training deep feedforward neural
  networks.
\newblock In {\em Proceedings of the thirteenth international conference on
  artificial intelligence and statistics}, pages 249--256, 2010.

\bibitem{goodfellow2016}
I.~Goodfellow, Y.~Bengio, and A.~Courville.
\newblock {\em Deep Learning}.
\newblock MIT Press, 2016.
\newblock \url{http://www.deeplearningbook.org}.

\bibitem{han2015learning}
S.~Han, J.~Pool, J.~Tran, and W.~Dally.
\newblock Learning both weights and connections for efficient neural network.
\newblock In {\em Advances in neural information processing systems}, pages
  1135--1143, 2015.

\bibitem{jin2016deep}
X.~Jin, C.~Xu, J.~Feng, Y.~Wei, J.~Xiong, and S.~Yan.
\newblock Deep learning with s-shaped rectified linear activation units.
\newblock In {\em Thirtieth AAAI Conference on Artificial Intelligence}, 2016.

\bibitem{lapshyna2020sparse}
V.~Lapshyna.
\newblock Sparse artificial neural networks: Adaptive performance-based
  connectivity inspired by human-brain processes.
\newblock {B.S.} thesis, University of Twente, 2020.

\bibitem{lecun1990optimal}
Y.~LeCun, J.~S. Denker, and S.~A. Solla.
\newblock Optimal brain damage.
\newblock In {\em Advances in neural information processing systems}, pages
  598--605, 1990.

\bibitem{liu_2017}
D.~Liu.
\newblock A practical guide to relu, Nov 2017.

\bibitem{liu2019sparse}
S.~Liu, D.~C. Mocanu, A.~R.~R. Matavalam, Y.~Pei, and M.~Pechenizkiy.
\newblock Sparse evolutionary deep learning with over one million artificial
  neurons on commodity hardware.
\newblock {\em arXiv preprint arXiv:1901.09181}, 2019.

\bibitem{maksutov_2018}
R.~Maksutov.
\newblock Deep study of a not very deep neural network. part 2: Activation
  functions, May 2018.

\bibitem{mishkin2015all}
D.~Mishkin and J.~Matas.
\newblock All you need is a good init.
\newblock {\em arXiv preprint arXiv:1511.06422}, 2015.

\bibitem{dcmocanuphd}
D.~C. Mocanu.
\newblock {\em Network computations in artificial intelligence}.
\newblock PhD thesis, Technische Universiteit Eindhoven, 2017.

\bibitem{mocanu2018scalable}
D.~C. Mocanu, E.~Mocanu, P.~Stone, P.~H. Nguyen, M.~Gibescu, and A.~Liotta.
\newblock Scalable training of artificial neural networks with adaptive sparse
  connectivity inspired by network science.
\newblock {\em Nature communications}, 9(1):1--12, 2018.

\bibitem{mostafa2019parameter}
H.~Mostafa and X.~Wang.
\newblock Parameter efficient training of deep convolutional neural networks by
  dynamic sparse reparameterization.
\newblock {\em arXiv preprint arXiv:1902.05967}, 2019.

\bibitem{najafabadi2015deep}
M.~M. Najafabadi, F.~Villanustre, T.~M. Khoshgoftaar, N.~Seliya, R.~Wald, and
  E.~Muharemagic.
\newblock Deep learning applications and challenges in big data analytics.
\newblock {\em Journal of Big Data}, 2(1):1, 2015.

\bibitem{nwankpa2018activation}
C.~Nwankpa, W.~Ijomah, A.~Gachagan, and S.~Marshall.
\newblock Activation functions: Comparison of trends in practice and research
  for deep learning.
\newblock {\em arXiv preprint arXiv:1811.03378}, 2018.

\bibitem{ohn2019smooth}
I.~Ohn and Y.~Kim.
\newblock Smooth function approximation by deep neural networks with general
  activation functions.
\newblock {\em Entropy}, 21(7):627, 2019.

\bibitem{ramachandran2017searching}
P.~Ramachandran, B.~Zoph, and Q.~V. Le.
\newblock Searching for activation functions.
\newblock {\em arXiv preprint arXiv:1710.05941}, 2017.

\bibitem{ramchoun2016multilayer}
H.~Ramchoun, M.~A.~J. Idrissi, Y.~Ghanou, and M.~Ettaouil.
\newblock Multilayer perceptron: Architecture optimization and training.
\newblock {\em IJIMAI}, 4(1):26--30, 2016.

\bibitem{overfitting}
S.~Saxena.
\newblock Underfitting vs. overfitting (vs. best fitting) in machine learning,
  Feb 2020.

\bibitem{sharma2017understanding}
A.~Sharma.
\newblock Understanding activation functions in neural networks.
\newblock {\em Medium. com blog}, 30, 2017.

\end{thebibliography}


\clearpage
\onecolumn
\appendix
\section{Performance}
\label{sec:performance}

\begin{figure}[!htb]
\centering
\begin{subfigure}{.49\textwidth}
  \centering
  \includegraphics[width=\linewidth]{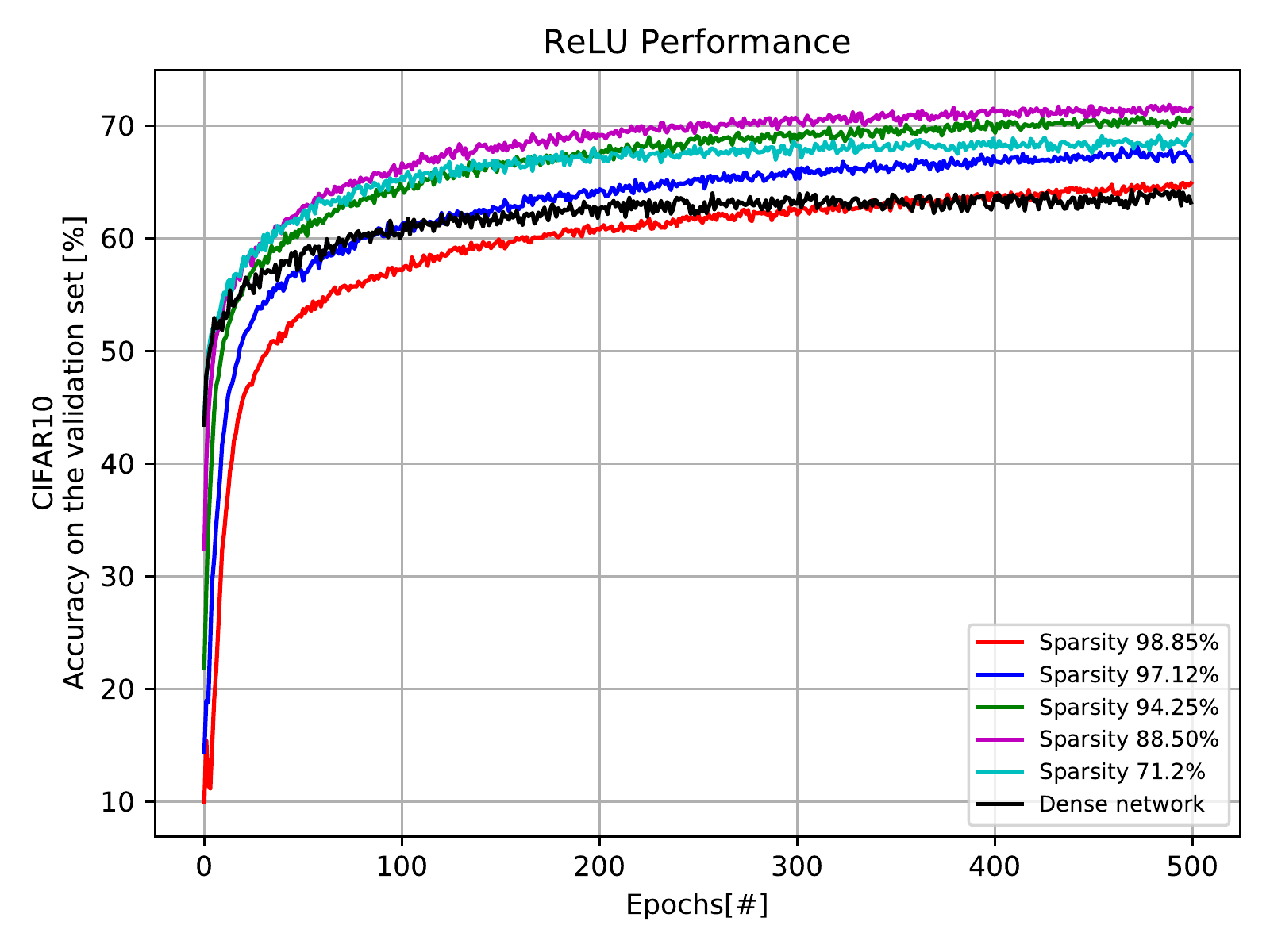}
  \caption{ReLU}
  \label{subfig:performance_relu}
\end{subfigure}%
\quad
\begin{subfigure}{.49\textwidth}
  \centering
  \includegraphics[width=\linewidth]{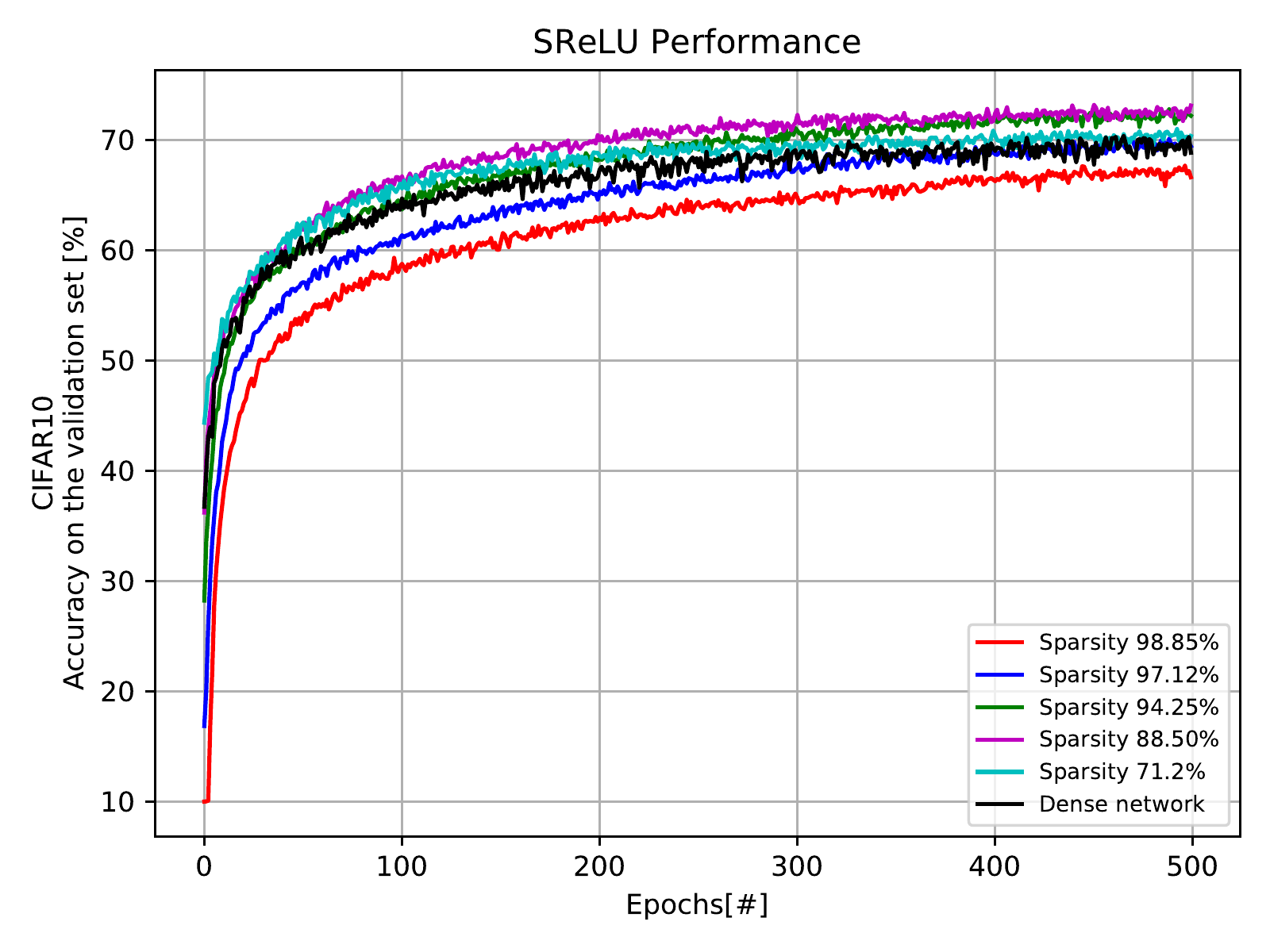}
  \caption{SReLU}
  \label{subfig:performance_srelu}
\end{subfigure}

\vskip\baselineskip

\begin{subfigure}{.49\textwidth}
  \centering
  \includegraphics[width=\linewidth]{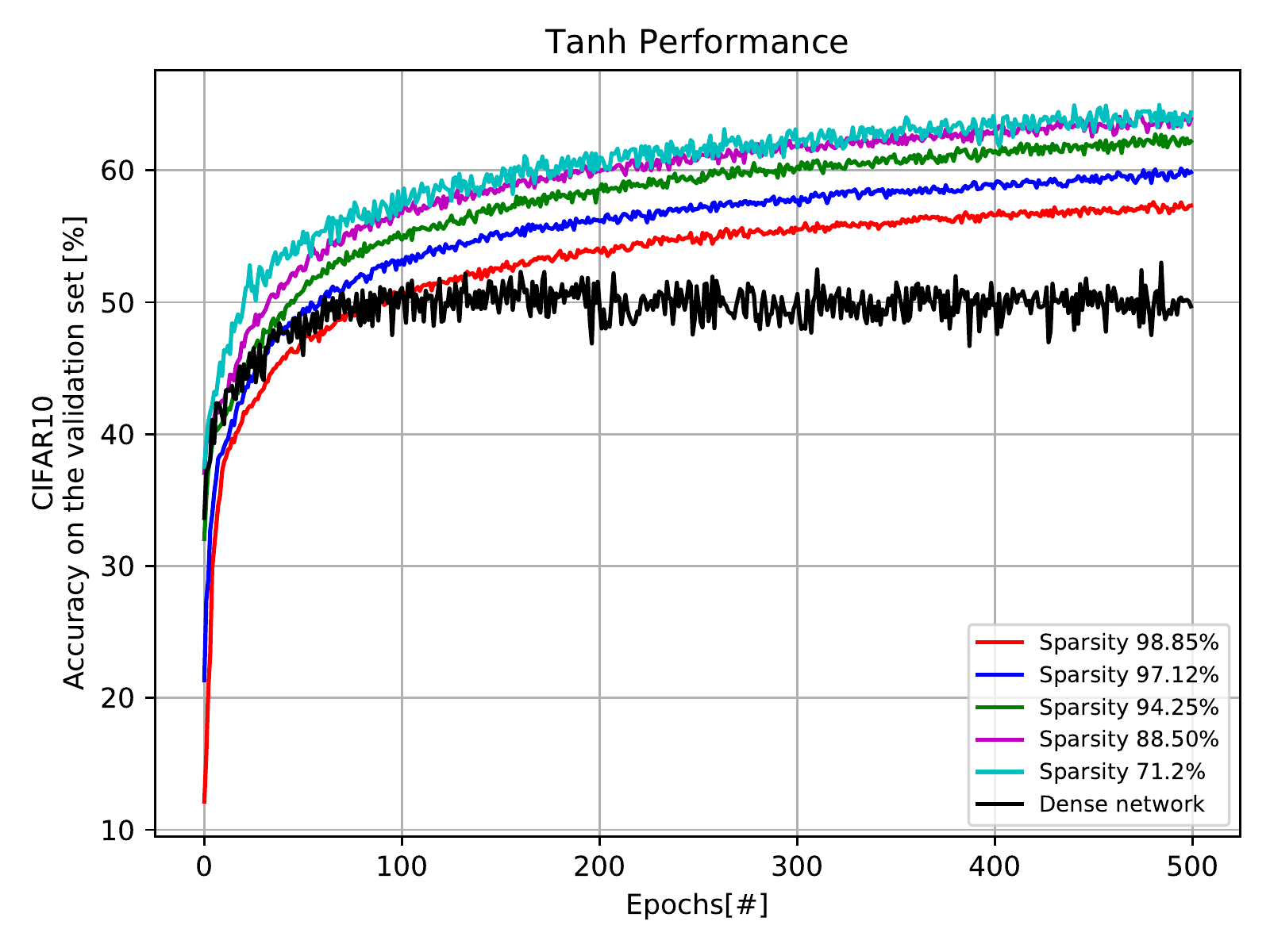}
  \caption{Tanh}
  \label{subfig:performance_tanh}
\end{subfigure}
\quad
\begin{subfigure}{.49\textwidth}
  \centering
  \includegraphics[width=\linewidth]{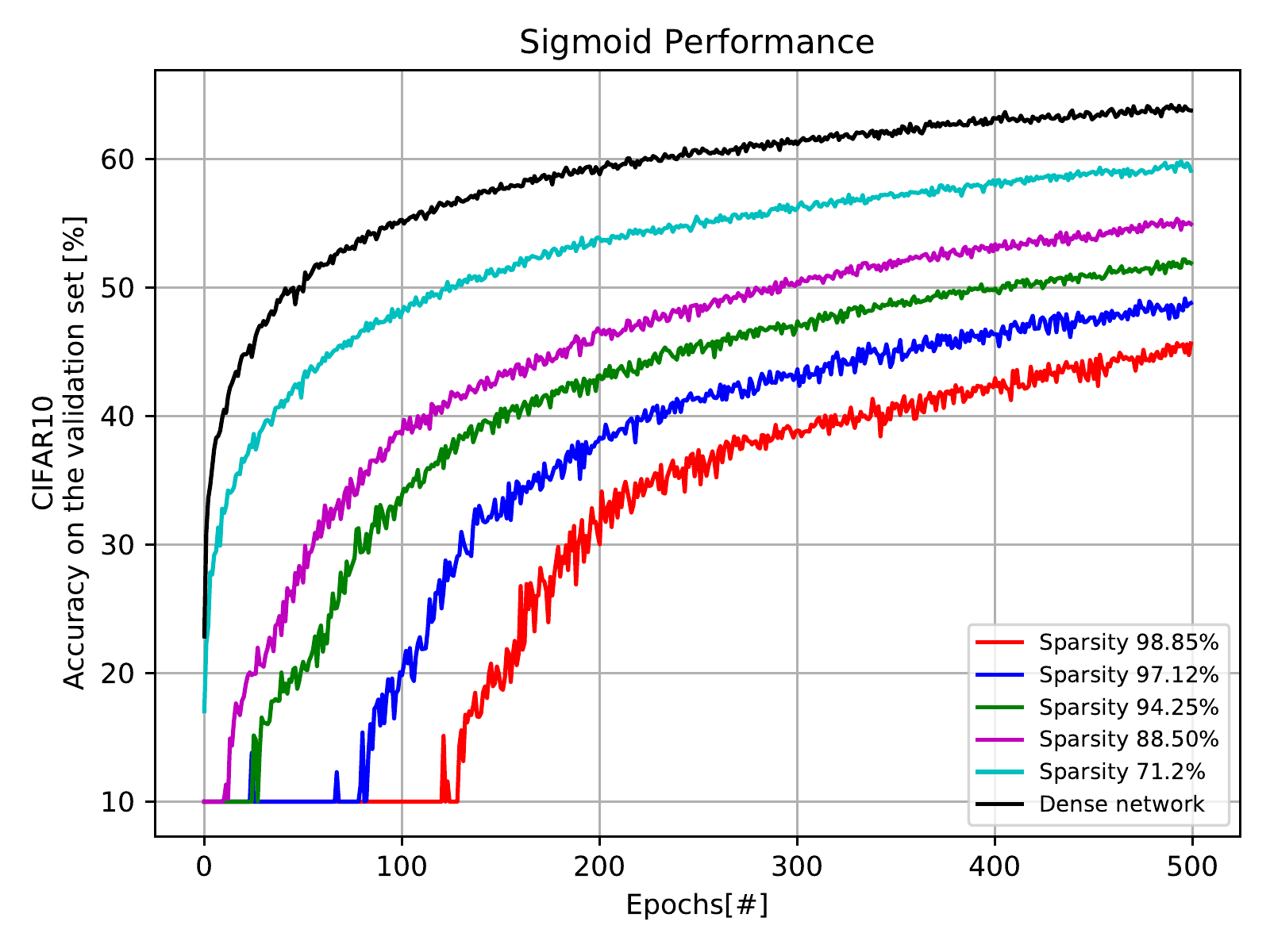}
  \caption{Sigmoid}
  \label{subfig:performance_sigmoid}
\end{subfigure}%

\vskip\baselineskip

\begin{subfigure}{.49\textwidth}
  \centering
  \includegraphics[width=\linewidth]{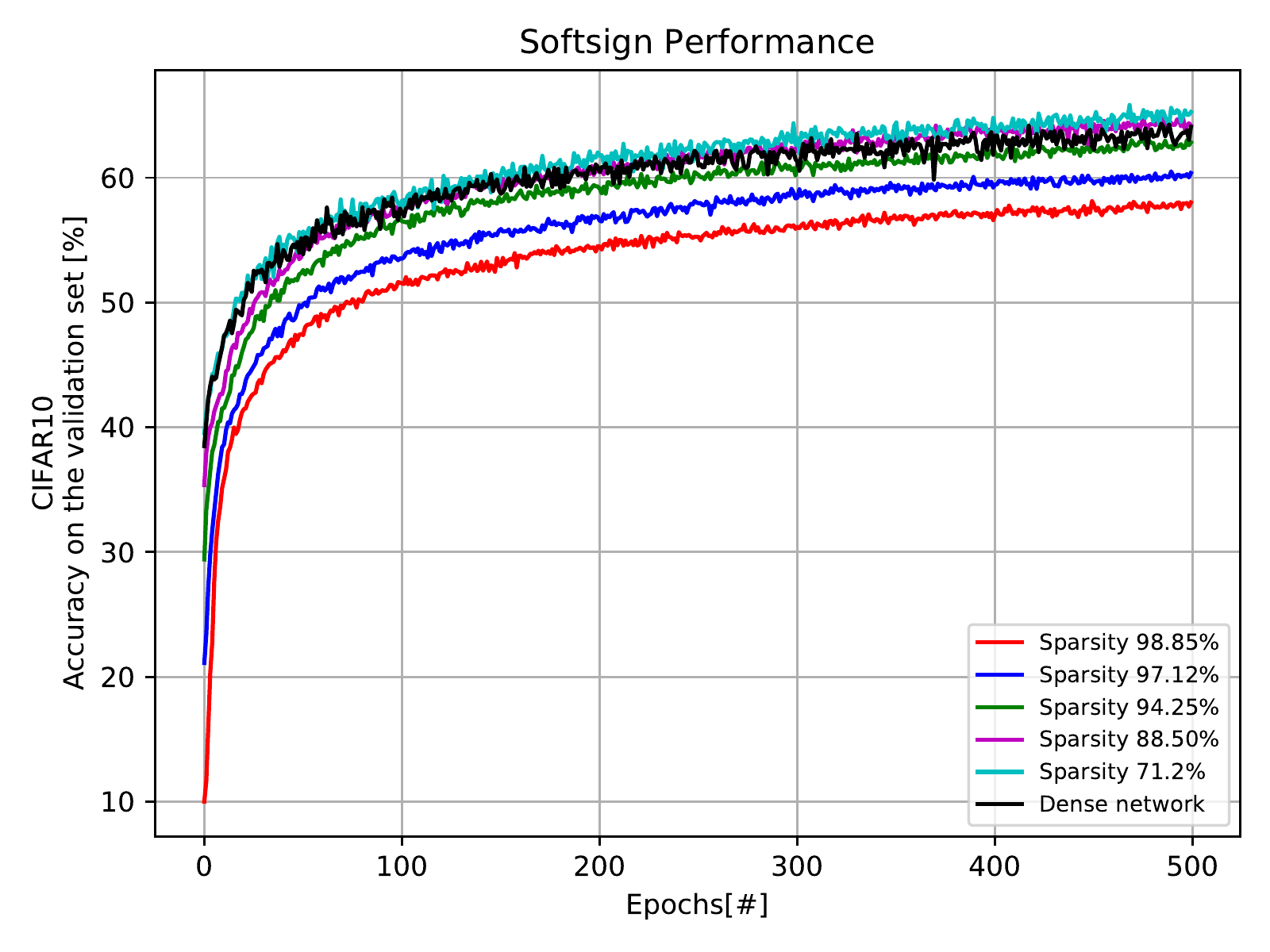}
  \caption{Softsign}
  \label{subfig:performance_softsign}
\end{subfigure}
\quad
\begin{subfigure}{.49\textwidth}
  \centering
  \includegraphics[width=\linewidth]{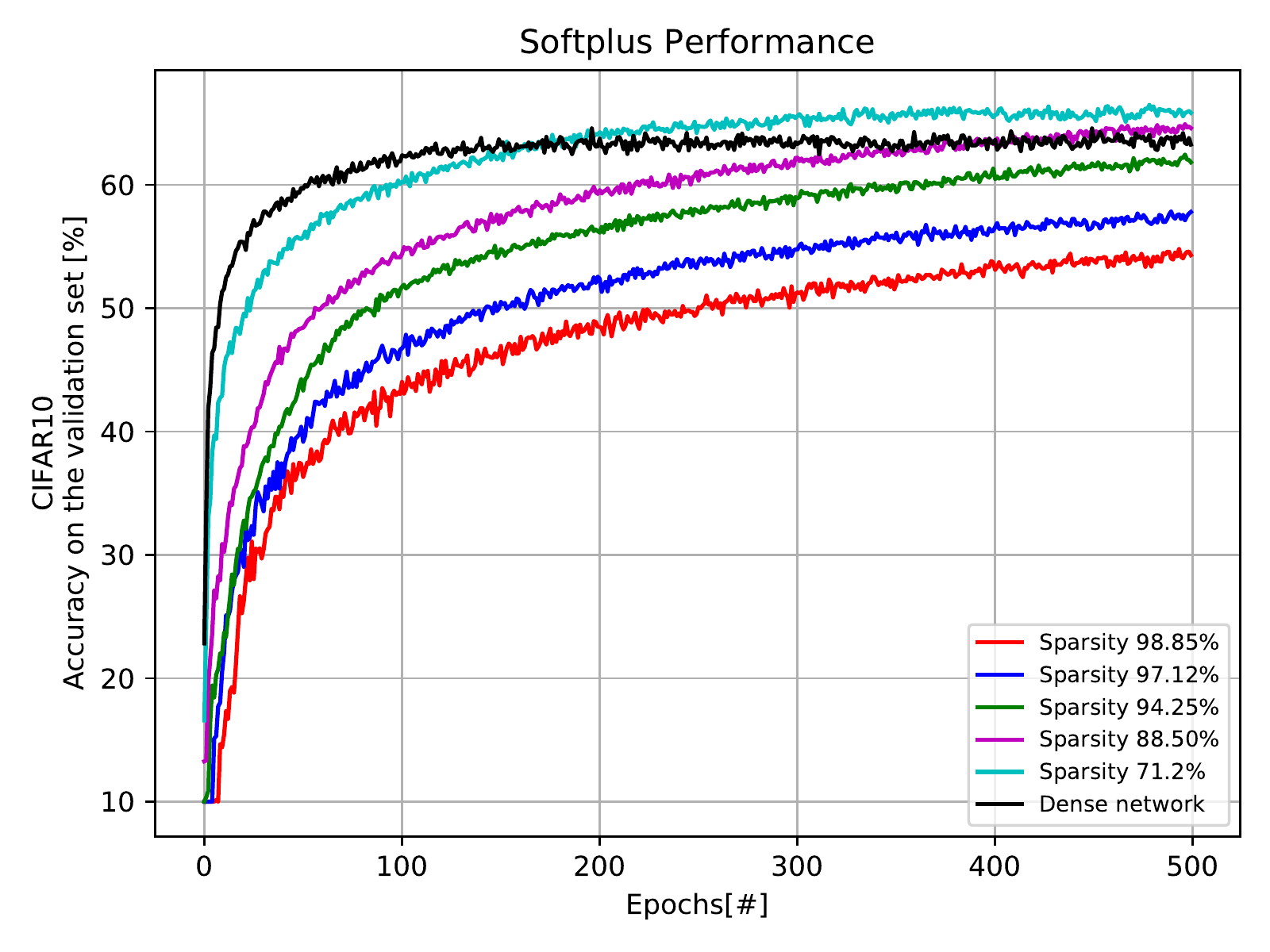}
  \caption{Softplus}
  \label{subfig:performance_softplus}
\end{subfigure}

\caption{Accuracy on the validation set per function.}
\label{fig:performance_per_function}
\end{figure}







\end{document}